\documentclass[review]{elsarticle}
\usepackage{lineno,hyperref}
\usepackage{amsmath,fancyhdr,amsthm,amssymb,amsfonts,version}
\usepackage{multirow}
\usepackage{bbm,version}
\usepackage{mathrsfs}
\usepackage[section]{placeins}
\usepackage{stmaryrd}
\usepackage{bm}
\usepackage{epstopdf}
\usepackage{subfig}
\usepackage{float}
\usepackage{subfloat}
\usepackage{graphicx}
\usepackage{verbatim}
\usepackage{diagbox}
\usepackage[noend]{algpseudocode}
\usepackage{algorithmicx,algorithm}
\include{pythonlisting}
\usepackage{threeparttable}
\usepackage{makecell}

\hypersetup{colorlinks=true, linkcolor=blue, citecolor=red}

\makeatletter
\newcommand\figcaption{\def\@captype{figure}\caption}
\newcommand\tabcaption{\def\@captype{table}\caption}
\makeatother

\numberwithin{equation}{section}

\begin{document}
	\graphicspath{ {./figure/} }
	
	\begin{frontmatter}
		
		\title{Operator Learning Enhanced Physics-informed Neural Networks for Solving Partial Differential Equations Characterized by Sharp Solutions}


            \author[mymainaddress1]{Bin Lin}
		\author[mymainaddress1]{Zhiping Mao\corref{mycorrespondingauthor}}
            \author[mymainaddress2]{Zhicheng Wang\corref{mycorrespondingauthor}}
		%
        \author[mymainaddress3]{George Em Karniadakis}
		
	\address[mymainaddress1]{School of Mathematical Sciences, Fujian Provincial Key Laboratory of Mathematical Modeling and High-Performance Scientific Computing, Xiamen University, Xiamen, 361005, China.}
        \address[mymainaddress2]{Laboratory of Ocean Energy Utilization of Ministry of Education, School of Energy and Power Engineering, Dalian University of Technology, Dalian, China.}
        \address[mymainaddress3]{Division of Applied Mathematics, Brown University, Providence, RI 02906, USA.}

           \cortext[mycorrespondingauthor]{Corresponding authors.
       Email addresses: zpmao@xmu.edu.cn (Z. Mao), zhicheng\_wang@dlut.edu.cn (Z. Wang)}

\begin{abstract}
Physics-informed Neural Networks (PINNs) have been shown as a promising approach for solving both forward and inverse problems of partial differential equations (PDEs). Meanwhile, the neural operator approach, including methods such as Deep Operator Network (DeepONet) and Fourier neural operator (FNO), has been introduced and extensively employed in approximating solution of PDEs. Nevertheless, to solve problems consisting of sharp solutions poses a significant challenge when employing these two approaches. To address this issue, we propose in this work a novel framework termed Operator Learning Enhanced Physics-informed Neural Networks (OL-PINN). Initially, we utilize DeepONet to learn the solution operator for a set of smooth problems relevant to the PDEs characterized by sharp solutions. Subsequently, we integrate the pre-trained DeepONet with PINN to resolve the target sharp solution problem. We showcase the efficacy of OL-PINN by successfully addressing various problems, such as the nonlinear diffusion-reaction equation, the Burgers equation and the incompressible Navier-Stokes equation at high Reynolds number. Compared with the vanilla PINN, the proposed method requires only a small number of residual points to achieve a strong generalization capability. Moreover, it substantially enhances accuracy, while also ensuring a robust training process. Furthermore, OL-PINN inherits the advantage of PINN for solving inverse problems. To this end, we apply the OL-PINN approach for solving problems with only partial boundary conditions, which usually cannot be solved by the classical numerical methods, showing its capacity in solving ill-posed problems and consequently more complex inverse problems. 
\end{abstract}
		
		
\end{frontmatter}

\section{Introduction}
Deep learning has achieved remarkable success in classical artificial intelligence tasks, such as image recognition, object detection, and natural language processing. Furthermore, deep learning has expanded its reach into scientific computing, where it has been proven to be a powerful tool for solving and learning partial differential equations (PDEs).  In addition to classical numerical methods, deep learning techniques offer an alternative approach by leveraging neural networks to directly approximate PDE solutions from data. This approach, often referred as ``physics-informed deep learning" or ``deep learning for PDEs" has shown great promise.  
We refer to \cite{raissi2019physics,yu2018deep, sirignano2018dgm,karniadakis2021physics, chen1995universal, lu2022multifidelity,lu2021deepxde,meng2020composite} and references therein.  

One typical and remarkable methodology is the so called Physics-informed Neural Networks (PINNs) proposed by Karniadakis et al \cite{raissi2019physics, lu2019deeponet}.  
PINN offers an alternative approach by leveraging the power of neural networks to approximate the solution to the physics-based equations. PINNs have been shown that it is a powerful tool in solving PDEs, especially for the inverse problems and high-dimensional problems. It has been now successfully applied to a wide range of scientific and engineering problems, such as fluid dynamics~\cite{cai2021physics}, solid mechanics~\cite{haghighat2021physics}, heat transfer~\cite{cai2021physics}, high-speed flows~\cite{mao2021deepm, lv2023deepstsnet}, and material science~\cite{zhang2020physics} due to the fact that PINNs have several advantages such as easy coding and conveniently handling complex geometries and particularly suitable in dealing with inverse problems.

However, when PINN is used to predict the solution of sharp spatio-temporal transitions with a small numbers of collocation points, it may produce a large generalization error. 
It has been pointed out that it is difficult to use neural networks to capture sharp fluctuations of functions~\cite{rahaman2019spectral}.
To resolve this issue, in cases where the precise locations of sharp features are identified,  adding a large number of residual points around these sharp locations is used to improve the generalization ability~\cite{mao2020physics, jagtap2022physics, mao2021deepm}. 
Several other efforts have been proposed using adaptivity, for instance, (1) adaptive sampling method~\cite{lu2019deeponet, mao2023physics, wu2023comprehensive, gao2023failure, han2022residual, hanna2022residual}, (2) self-adaptive PINN \cite{mcclenny2020self}, (3) weighted PINN for Euler \cite{xiong2022gradient}, (4) adaptive activation function \cite{jagtap2020adaptive,jagtap2020locally, jagtap2022deep}. All aforementioned adaptive extensions of PINN are usually computationally expensive and/or unstable. PDEs in the weak sense are also used for the singular problems to improve the approximation ability of neural network-based method but it cannot help the training efficiency~\cite{zang2020weak, bao2020numerical, de2022weak, yu2018deep}. 
Also, to solve the singularly perturbed problems, a deformation of the traditional PINN based on singular perturbation theory is proposed in \cite{arzani2023theory}, and the deep operator approximation with Shishkin mesh points is proposed in \cite{du2023approximation}.

As we know, even for the classical numerical methods, it is usually challenge to solve problems whose solutions exhibit sharpness/shock such as the hyperbolic conservation laws~\cite{dafermos2005hyperbolic, harten1983upstream}, incompressible flows with high Reynolds numbers~\cite{temam2001navier} and so on. 
However, these ``low viscous (high Reynolds number)" solutions usually can be asymptotically approached by ``high viscous (low Reynolds number)" solutions.
For instance, in order to \emph{analyze} or \emph{approximate} nonsmooth solutions of a hyperbolic equation, a strategy is to analyze the viscous smooth solutions and then let the viscosity goes to zero, or to solve a corresponding viscous problem with a vanishing viscosity.
For example, consider the following hyperbolic equations  
	\begin{equation}\label{eq:hyper}
		u_t + \nabla f(u) = 0.
	\end{equation}
It is well-known that the solutions of the above equation are usually discontinuous and weak solutions are not unique. To obtain the the uniqueness of the above equation, an artificial viscosity is introduced, namely, we consider the following corresponding viscous problem 
	\begin{equation*}\label{eq:hyper:vis}
		u_t + \nabla f(u) = \nu \Delta u,
  \end{equation*}
and then we obtain an unique weak entropy solution satisfying the vanishing viscosity for the hyperbolic equation \eqref{eq:hyper} as $\nu$ goes to zero.
On the other hand, from the computational point of view, to design a stable algorithm for numerically solving the steep gradient problems, one of the popular techniques is to add an artificial viscosity to the original problems, for instance, spectral vanishing viscosity~\cite{karamanos2000spectral}, entropy viscosity method~\cite{guermond2011entropy,WANG2019108832}.
This kind of idea has also been used with PINN. \citet{he2023artificial} and \citet{wang2023solution} proposed an artificial viscosity augmented PINN to resolve the solution multiplicity issue, while 
Coutinho et al developed an adaptive locaized artificial viscosity PINN~\cite{coutinho2023physics}.
In \cite{de2022wpinns}, the entropy condition is added in the loss function to obtain the unique solution. Trask et al developed thermodynamically consistent PINNs for hyperbolic systems by considering the entropy condition~\cite{patel2022thermodynamically}. Dong et al proposed a variant of PINN using new forms of loss function for solving the hyperbolic PDEs~\cite{qian2023physics}.

Motivated by this, we propose herein an operator learning enhanced PINN (OL-PINN) to solve this kind of singular problems. To this end, we aim to first solve a class of smooth problems by using the deep operator learning, and we adopt the deep operator neural networks (DeepONets) proposed by Lu et al~\cite{lu2019deeponet}. Then, we combine the pretrained DeepONet with PINNs to solve the corresponding singular problem. The details will be given in the next section.
\emph{The basic idea is to use the pretrained operator with a slight modification as an additional regularization to improve the stability and efficiency of the training of PINNs.}
In principle, this coincides with the methodology of metalearning with fine tuning. 
We also point out that in the case of using the artificial viscosity, the resulted solutions differ from those of the original PDES; to this end, it is important to recommend an appropriate quantitative artificial viscosity based on empirical evidence. However, in the present work, we solve the \emph{original problems} without using an artificial viscosity. Instead, we use the extrapolation of the viscous solution operator as an additional regularization. 
Some other works related with PINN and operator learning can be found in \cite{wang2021learning, li2021physics, goswami2023physics, zhu2023reliable, hao2023instability}.
Our method has the following advantages:
\begin{itemize}
    \item The \emph{generalization ability is significantly improved} compared with the vanilla PINN. This is to say we obtain a satisfactory test accuracy with small number of residual points.
    \item Also, the use of a small number of residual points \emph{improves the efficiency} of the training of OL-PINN since computing the derivatives with respect to the input is extremely time consuming by using auto differentiation. 
    \item The training process of OL-PINN is \emph{robust}. 
\end{itemize}

Furthermore, 
compared with the classical numerical methods, another \emph{big advantage} of the present method is that the present method possesses the ability of solving problem with insufficient initial-boundary conditions. In certain fluid dynamics applications, situations arise where obtaining adequate boundary conditions for mathematical models becomes challenging, if not outright impossible. Therefore, we also consider in this work another class of problems, i.e., the ill-posed problems with insufficient boundary conditions. Numerical results show that we still obtain good results by using the present framework for this kind of ill-posed problems. 

The rest of the paper is organized as follows. PINNs and DeepONets as well as the proposed OL-PINN are introduced in Section \ref{sec:method}. Subsequently, we show the numerical results for the nonlinear diffusion-reaction equation, Burgers equations and the Navier-Stokes equations to demonstrate the effectiveness of the present method in Section \ref{sec:numerics}. A summary with a short discussion are given in Section \ref{sec:summary}.

\section{Methodology}\label{sec:method}
In this section, we introduce the operator learning enhanced PINNs in this section. In the first two subsections, we introduce the frameworks of PINNs and DeepONets, then we propose the operator learning enhanced PINNs. 

\subsection{PINN:Physics-informed neural network}
\begin{figure}[htbp]
		\centering
		\includegraphics[scale=0.5]{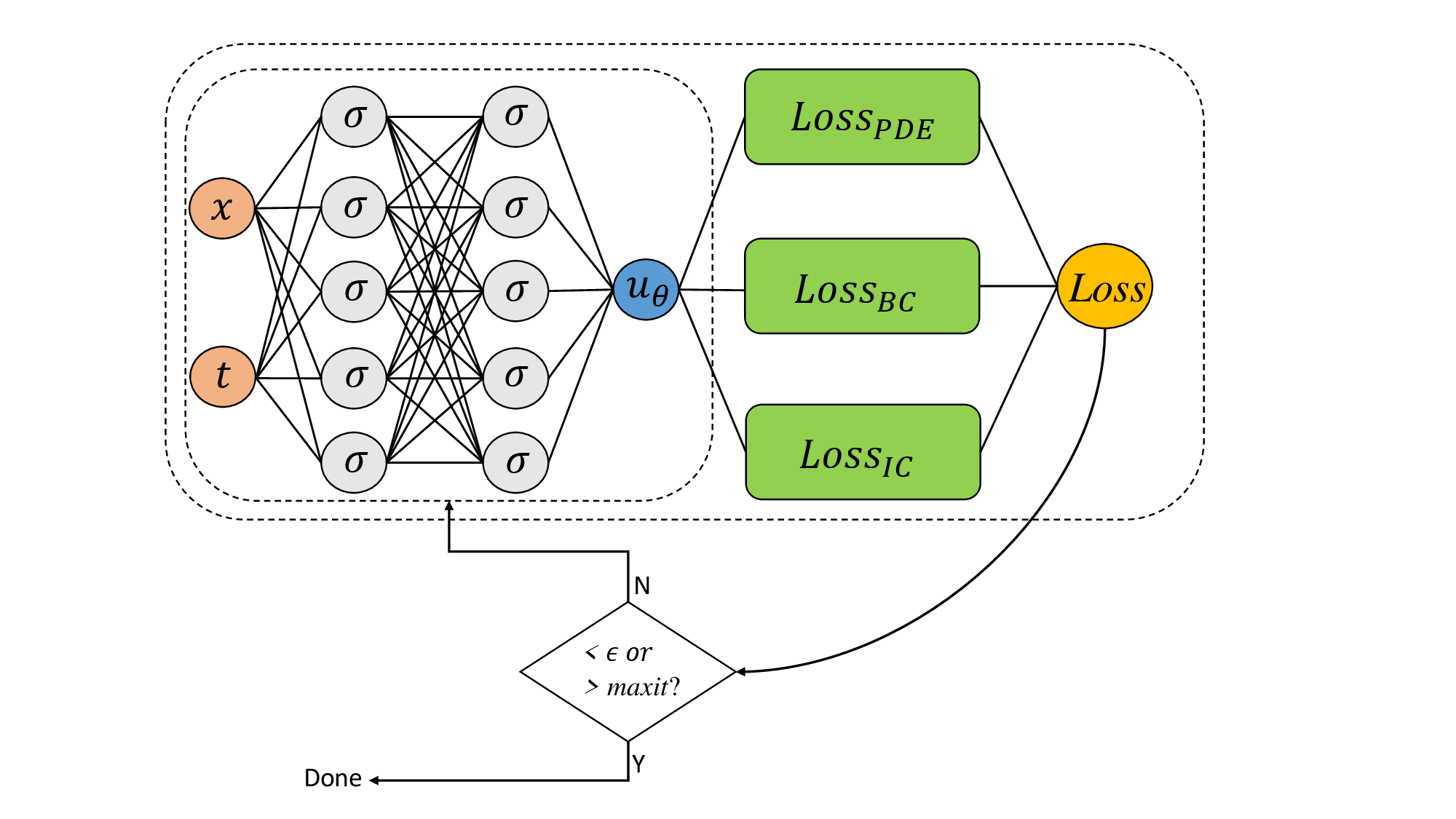}
		\caption{Schematic of PINNs for solving PDEs.}
		\label{PINN}
	\end{figure}

We begin by briefly introducing PINNs for solving PDEs. Consider the following nonlinear PDEs:
\begin{equation}\label{pde}
    \begin{aligned}
		& \mathcal{N}[u(\boldsymbol{x}, t)]=0, \quad \boldsymbol{x} \in \Omega, t \in[0, T], \\
		& u(\boldsymbol{x}, t)=g(\boldsymbol{x}, t),\quad \boldsymbol{x} \in \partial \Omega, t \in[0, T],\\
		& u(\boldsymbol{x}, 0)=h(\boldsymbol{x}),\quad \boldsymbol{x} \in \Omega.
	\end{aligned}
\end{equation}
For the neural network-based deep learning, the solution $u(\boldsymbol{x}, t)$ is approximated by a neural network denoted by $u_{\theta}(\boldsymbol{x},t)$, which takes $(\boldsymbol{x}, t)$ as the input (see Figure \ref{PINN}), where $\theta$ represents the trainable parameters. In the original PINN framework, to solve the above equation, we minimize the following loss function:
\begin{align}\label{loss:pinn}
  Loss = Loss_{PDE} + Loss_{Data},
\end{align}
where $Loss_{Data}$ is the mismatch correspongding to the prescribed data. For instance, we have 
$$Loss_{Data}= Loss_{BC} + Loss_{IC}$$ 
when solving the forward problems, where 
\begin{align*}
		& Loss_{PDE}=\frac{1}{N_r} \sum_{i=1}^{N_r} \left| \mathcal{N}[u_{\theta}(\boldsymbol{x}_i^r, t_i^r)]\right|^2, \\
		& Loss_{BC}=\frac{1}{N_b} \sum_{i=1}^{N_b}\left|u_{\theta}\left(\boldsymbol{x}_i^b, t_i^b  \right)-g(\boldsymbol{x}_i^b, t_i^b)\right|^2, \\
		& Loss_{IC}=\frac{1}{N_0} \sum_{i=1}^{N_0}\left|u_{\theta}\left(\boldsymbol{x}_i^0, 0 \right)-h(\boldsymbol{x}_i^0, 0)\right|^2
\end{align*}
are losses associated with the equation, boundary and initial conditions, respectively. 
Here $(\boldsymbol{x}_i^r, t_i^r)$ are the residual points sampled over the domain of interest for the PDE, $(\boldsymbol{x}_i^b, t_i^b)$ and $(\boldsymbol{x}_i^b, 0)$ are two sets of points corresponding to the boundary condition and initial condition, respectively. $N_r, ~ N_b$ and $N_0$ are the total numbers of residual points, boundary points and initial points, respectively. Then, an optimizer (e.g. SGD, Adam) is applied to solve the optimization problem to obtain the solution. The schematic of a PINN is shown in Figure \ref{PINN}.
	

\subsection{DeepONet: Deep operator neural network}
\begin{figure}[htbp]
\centering
\includegraphics[scale=0.4]{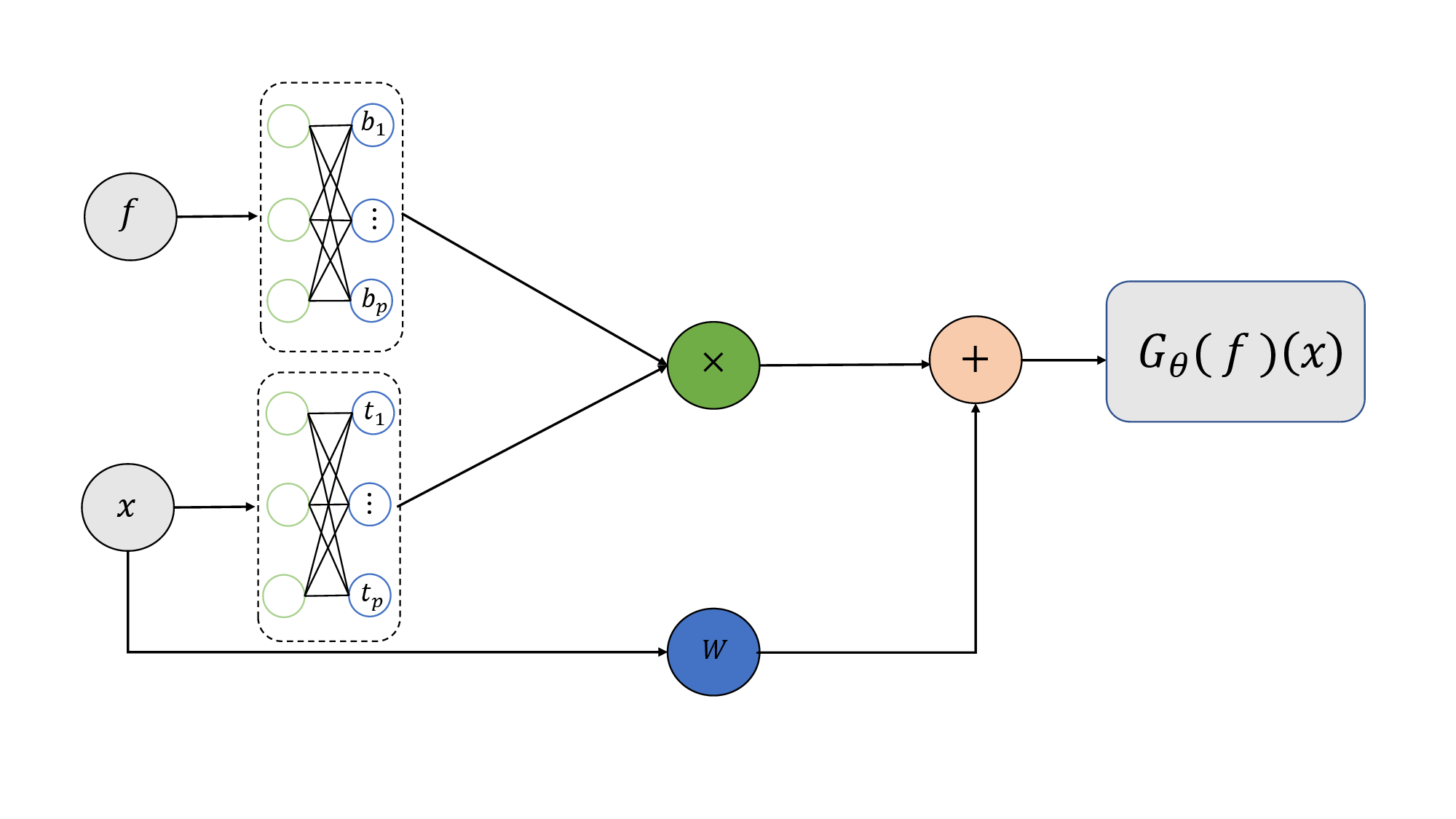}
\caption{Schematic of DeepONet.}
\label{deeponet}
\end{figure}

Neural networks exhibit not only a capacity for the universal approximation of continuous functions but also possess the ability to universally approximate nonlinear continuous operators~\cite {chen1995universal}. 
In view of the universal approximation theory, Lu et al proposed a deep operator neural network (DeepONet), which learns the mapping between two functional spaces. Another popular operator learning methodology is the so-called Fourier neural operator (FNO), which is more efficient in training but not good at dealing with complex geometry problems. The operator learning is a quite efficient and powerful tool for solving a class of similar equations (e.g. parameterized PDEs).
Specifically, the DeepOnet is trained offline and can make predictions online without further training.
Now note that in the current work, we need to solve a class of smooth problems, and this can be accomplished by using the operator learning that expresses operator regression.
In the present work we adopt DeepOnet for its generality. We also point out here that we make a slight modification to the architecture of the original unstack DeepONet by adding a bias to the output. Numerical experiments show that the modified DeepONet has better performance in most cases. The description of the architecture of the DeepONet is given in the following and a schematic is shown in Figure \ref{deeponet}.

The operator learning is used to approach the following operator 
$$\mathit{G}:f\rightarrow\mathit{G}(f).$$ 
And we use a DeepONet to represent the solution operator denoted by $G_{{\theta}}({f})$, where $\theta$ is the parameters to be trained. A DeepONet consists of two sub-networks, a branch net and a trunk net. The branch net takes the discrete values of the input functions, i.e., $f_i^k, i=1, \ldots , m,\; k =1,\ldots, N_{sample}$, as input, while the trunk network and the linear transformation take the locations $x_j, j=1, \ldots,n$ as input. Here $N_{sample}$ is the number of training samples. The output of the DeepONet is given by 
	\begin{align}
		\mathit{G}_{\theta}(f)(\boldsymbol{x})=\sum_{k=1}^p b_k(f) t_k(\boldsymbol{x})+\mathit{W}(\boldsymbol{x}),
	\end{align}
where  $\left\{b_1, b_2, \ldots, b_p\right\}$ and $\left\{t_1, t_2, \ldots, t_p\right\}$ are the outputs of the branch net and the trunk net, respectively. Here we add $\mathit{W}(\boldsymbol{x})$, is a linear transformation with respect to the input $\boldsymbol{x}$, to the inner product of the branch net and trunk net as a bias. 

\subsection{OL-PINN: Operator learning enhanced PINN}
Now let us introduce the main architecture in this work, i.e., the architecture of the OL-PINN for solving PDEs.

Without a prior knowledge of the locations of the sharpness, as mentioned previously, training PINNs becomes particularly challenging when dealing with singular problems and a limited number of residual points. In these cases, it is much more difficult to train the neural networks, i.e., solve the nonconvex optimization problems, than the cases of solving smooth problems.
Therefore, besides the physical constrains characterized by PDEs, it is better to have more regularization. In this work, we shall use the extrapolation of the pretrained DeepONet developed for the corresponding smooth problems as an additional regularization. 

\begin{figure}[htbp]
		\centering
		\includegraphics[scale=0.45]{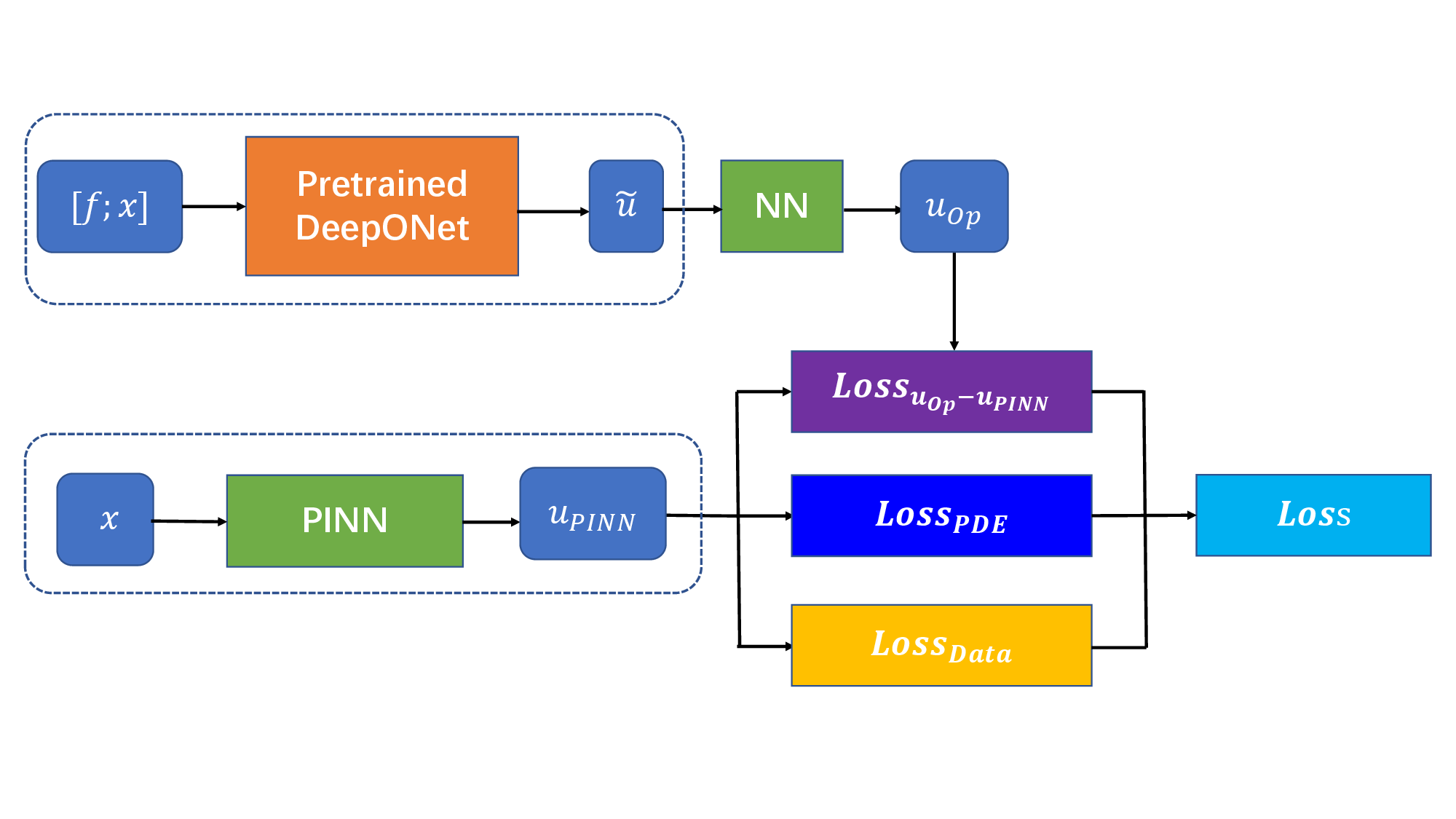}
		\caption{Schematic of OL-PINN. $u_{PINN}$ takes $x$ as input and the output would be used to compute the residual of the equation, while the output $u_{Op}$ takes the prediction of the pretrained DeepONet as input. The difference between these two outputs is used as an additional loss.}
		\label{PINN+onet}
	\end{figure}

To this end, we propose the architecture of OL-PINN (see Figure \ref{PINN+onet}) as follows. We construct two neural networks that to be trained. The first one inherits the architecture of PINN, which takes the location $x$ as input and ouputs $u_{PINN}$. For the second one, we take the prediction of the pretrained DeepONet (i.e., $\tilde{u}$ in Figure \ref{PINN+onet}) as input and output $u_{Op}$ and adopt the architecture shown in Figure \ref{deeponet}.
The loss function consists of three components given as follows:
\begin{align}\label{loss:ol-pinn}
		Loss = w_1 Loss_{PDE}  + w_2 Loss_{Data} + w_3 Loss_{u_{Op}-u_{PINN}},
\end{align}
where $w_1, w_2, w_3$ are prescribed weights. The first two are the losses associated with the PDE and the Data (initial and boundary conditions), which is essentially the same as the ones of PINNs given by \eqref{loss:pinn}. To \emph{avoid manually tuning the weights} for the PDE residual and the IC/BCs, we always set $\omega_1 = \omega_2$. Then, we take the mean square error of the mismatch between the outputs of the two neural networks as an additional regularization in the training, namely, the third component of the loss is given by 
\begin{align}\label{loss:pinn-op}
	Loss_{u_{Op}-u_{PINN}}=\frac{1}{N_c} \sum_{i=1}^{N_c}\left|{u_{Op}}\left(x_i^{c} \right)-u_{PINN}\left(x_i^{c} \right)\right|^2, 
\end{align}
where $x_i^c, i = 1,\ldots, N_c$ are the collocation points corresponding to the matching of the two networks (the third term of the loss \eqref{loss:ol-pinn}) with $N_c$ is the total number. 
We point out here that we also include points located at $t=0$ or on the boundaries in the collocation point set $\{x_i^c\}_{i=1}^{N_c}$.
We remark here that \emph{using a large value of $N_c$ would not affect the efficiency} since we do not need to compute any derivatives when computing the third term of \eqref{loss:ol-pinn}. On the other hand, we should \emph{use as few as possible residual points} due to the fact that computing derivatives of the neural network function using auto differentiation is very time consuming. The algorithm is given in Algorithm \ref{alg:ol}.

The operator learning excels in delivering highly accurate predictions when tasked with interpolation scenarios, yet they often fall short in producing favorable outcomes when confronted with extrapolation scenarios. Reliable extrapolations can be accomplished either using data or physics~\cite{mao2021deepm, zhu2023reliable}. The above methodology can also be interpreted as obtaining reliable extrapolations of DeepONets with the help of PINNs. 


	\begin{algorithm}[htbp]
		\caption{OL-PINN} %
		\hspace*{0.02in} {\bf Input:} 
		Pre-trained Deeponet, function $f$, residual and collocation points $\boldsymbol{x}^r,\boldsymbol{x}^c$, initial and boundary conditions $h(\bm{x}),\;g(\bm{x},t)$. \\
		\hspace*{0.02in} {\bf Output:} 
		Predictions: $u_{O p},\; u_{PINN}$
		
		\begin{algorithmic}[1]
			\State Construct two neural networks: NN and PINN. 
			
			
			\State Calculate $Loss_{PDE},~ Loss_{Data}$ and $Loss_{u_{Op}-u_{PINN}}$ to get the total loss. 
			
			\State Train the neural networks NN and PINN using the Adam optimizer.
		\end{algorithmic}
  \label{alg:ol}
	\end{algorithm}

\section{Numerical examples and results}\label{sec:numerics}
In this section, we present several numerical examples to illustrate the effectiveness of the present method for the nonlinear diffusion-reaction problem, Burgers equation, Navier-Stokes equation. Moreover, we solve the corresponding ill-posed problems for each equation by considering using partial boundary conditions.
We point out here that if not specified, the parameters used for each cases are presented in \ref{sec:apd:parameters}.

\subsection{Example 1: One-dimensional nonlinear diffusion-reaction equation}\label{section:onedim}

We begin by considering a smooth problem, i.e., the following one-dimensional nonlinear diffusion-reaction equation:  
\begin{equation}
\left\{ 
\begin{aligned}
  &-\Delta u+u^3-u = f, \;x \in(-1,1), \\
  &u(\pm 1) =0.
\end{aligned}
\right.
\end{equation}
We set the exact solution of the above equation to be $u(x)=a\sin \pi x$, where $a$ is a given parameter. Direct calculation yields $f(x) = a\left(\pi^2-1\right) \sin \pi x+(a\sin \pi x)^3$.  We first employ the DeepONet to learn the operator $a\rightarrow u(x)$ for $a \in [0,1]$ and solve the problem for the cases of $a = 5,10$. 

\begin{figure}[htbp]
		\centering
		\subfloat{\label{a5}
			\includegraphics[scale=0.33]{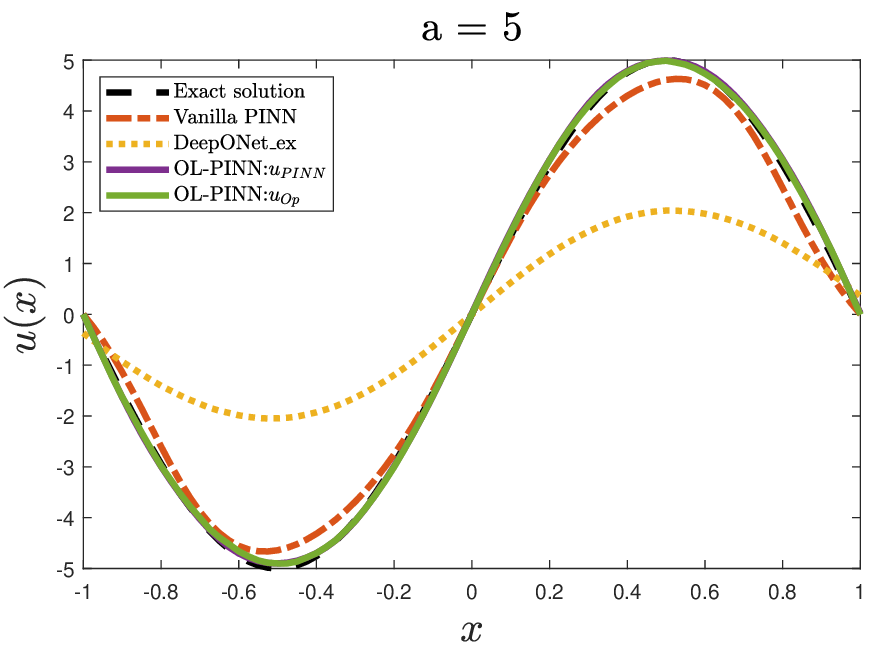}}
		\subfloat{\label{a10}
			\includegraphics[scale=0.33]{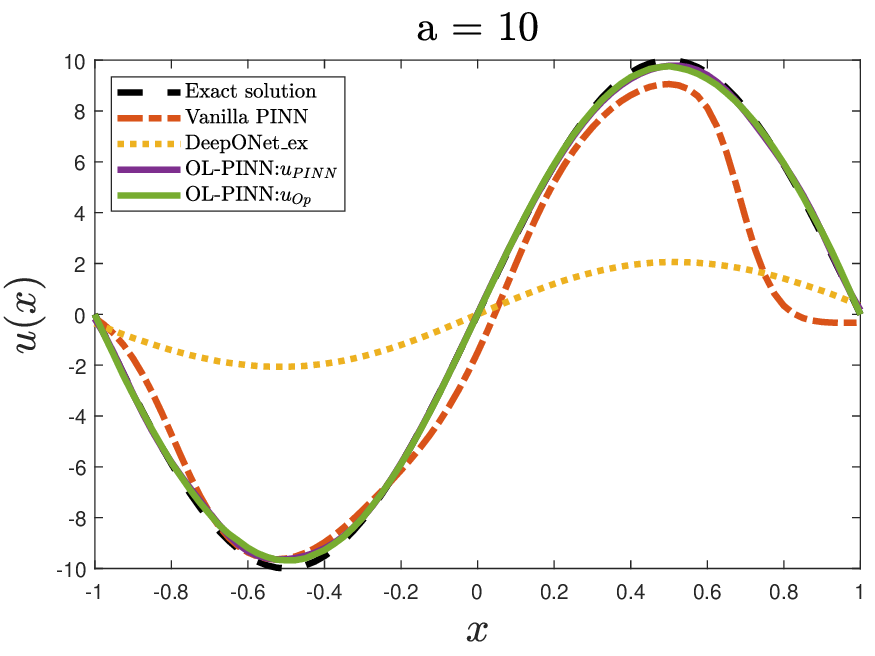}}
		\subfloat{\label{a_loss}
			\includegraphics[scale=0.33]{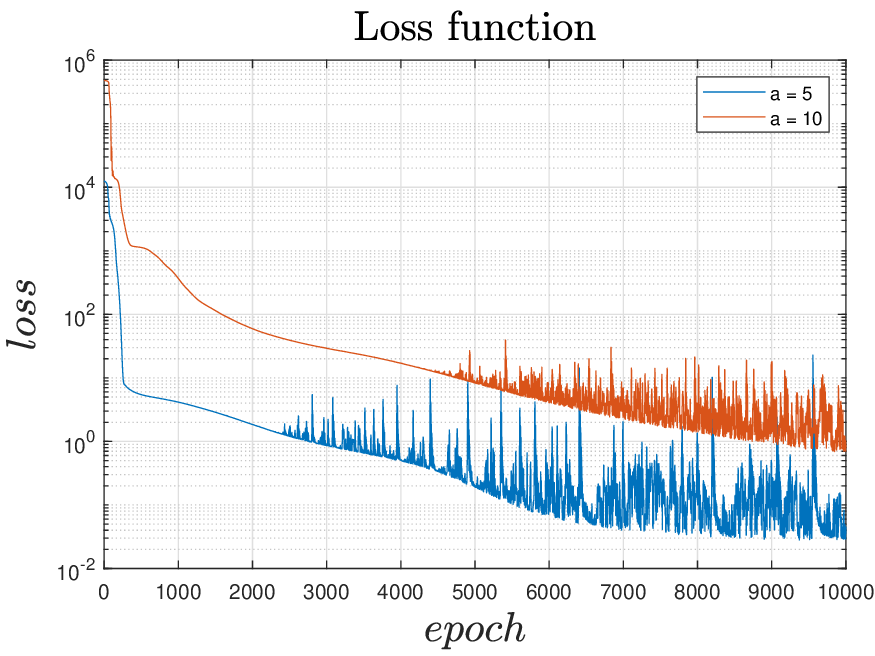}}
		\caption{Example 1: Comparison of the predictions with different models. Here we use only \emph{6 residual points}. Left: $a=5$, middle: $a=10$, right: Loss vs number of epochs.}
		\label{fig:Ex1}
	\end{figure}

To demonstrate the effectiveness of OL-PINN, we use only \emph{6 uniformly distributed residual points} in this example. We use the Adam optimizer with 10000 epochs and learning rate 0.001 for the training. The loss histories are given in the right plot of Figure \ref{fig:Ex1}. The results for $a = 5,10$ are shown in Figure \ref{fig:Ex1}. Observe that either the vanilla PINN or the extrapolation of DeepONet makes poor predictions. However, the predictions using OL-PINN are in good agreement with the exact solutions. We further show the mean and standard deviation of the relative $L^2$ error for different cases in Table \ref{tab:my_label} demonstrating that OL-PINN exhibits the capacity to effectively solve PDEs using only a small number of residual points.

	\begin{table}[htbp]
		\centering
		\begin{tabular}{|c|c|c|c|c|}
			\hline
            Model & Vanilla  PINN  & DeepONet\_{ex} & OL-PINN: $u_{PINN}$ & OL-PINN: $u_{Op}$ \\
			\hline  {a = 5 } & $ 10.21\pm 4.89  $ & $47.81  \pm 17.20 $& $\bm{1.86  \pm 0.29} $ & $1.88  \pm 0.32 $\\
			\hline  {a =10 } & $ 11.71 \pm   $ 4.74 & $67.70 \pm 16.80 $ & $\bm{1.83 \pm 0.45} $& $1.83 \pm 0.48 $\\

			\hline
		\end{tabular}\caption{Example 1: Mean and standard deviation of the relative $L^2(\%)$ test error for each model.}
		\label{tab:my_label}
	\end{table}

\subsection{Example 2: Burgers equation}
Now let us consider the one-dimensional time dependent Burgers equation 
\begin{equation}
\left\{ 
\begin{aligned} 
    &\partial_t u +u u_x =\nu \partial_{x}^2 u, \; x \in(-1,1), \,t \in(0,T], \\ 
    &u(x, 0) =-\sin\pi x, \; x \in(-1,1),\\
    &u(-1,t) = u(1,t) = 0, \;t \in[0,T],
\end{aligned}
\right.
\end{equation}
where $\nu$ is the viscous coefficient. In this example, we consider the following two cases:
\begin{itemize}
    \item Case I: We employ the present method to solve the above equation for both the \emph{viscid} case ($\nu = 0.001/\pi$) and the \emph{inviscid} case ($\nu=0$). The  pre-trained DeepONet is used to solve the problems for $\nu \in [0.02/\pi, 0.06/\pi]$. In this case, we set $t\in[0.0.9]$.
    \item Case II: We employ the present method to solve the above equation for $t\in [0.6, 0.8]$ with fixed value of $\nu = 0.008/\pi$. The pre-trained DeepONet is used to solve the problems for $t \in [0, 0.6]$.
\end{itemize}

\subsubsection{Case I: Prediction for unseen $\nu$}\label{section:caseI}
As we know, the solution of the above Burgers equation becomes sharper as $\nu$ gets smaller, and there exists a shock when $\nu = 0$ and $t$ is sufficiently large. Therefore, we train the DeepONet using the problems with smooth solutions (i.e., $\nu \in [0.02/\pi, 0.06/\pi]$) and then solve the problems with sharp solutions (i.e., $\nu = 0.001/\pi$ and $\nu = 0$).

\begin{figure}[htbp]
		\centering
		\subfloat[\label{fig5:a}$\nu = 0.001/\pi$]{
			\includegraphics[scale=0.45]{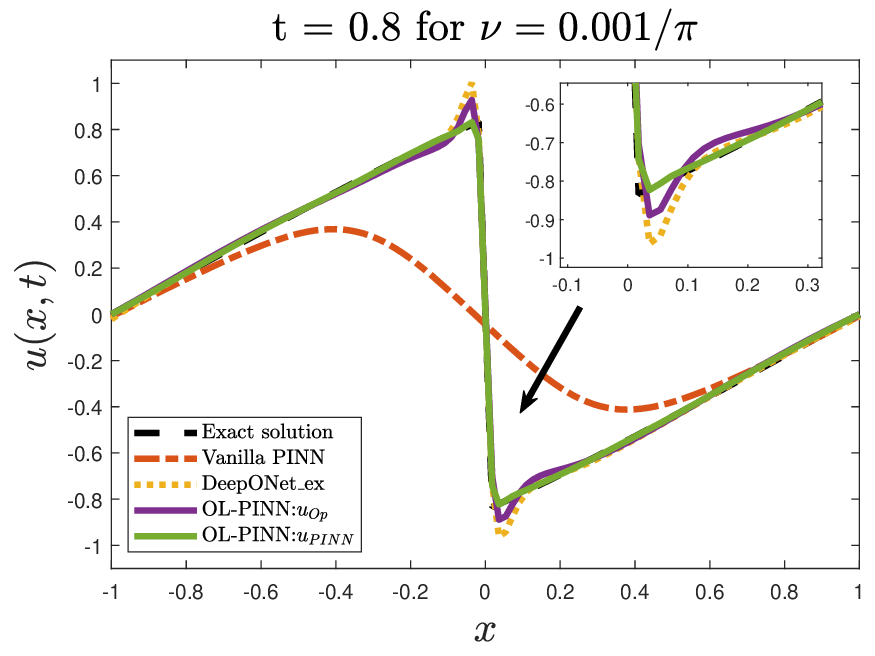}}
		\subfloat[\label{fig5:b}$\nu = 0$]{
			\includegraphics[scale=0.45]{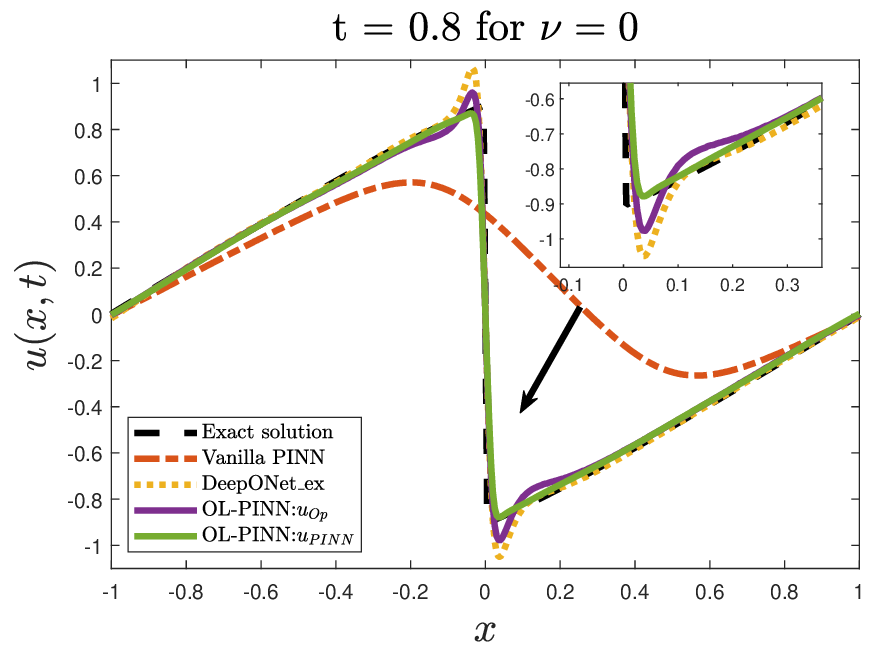}}
		\quad
		\subfloat[\label{fig5:c}$\nu = 0.001/\pi$ without BCs]{
			\includegraphics[scale=0.45]{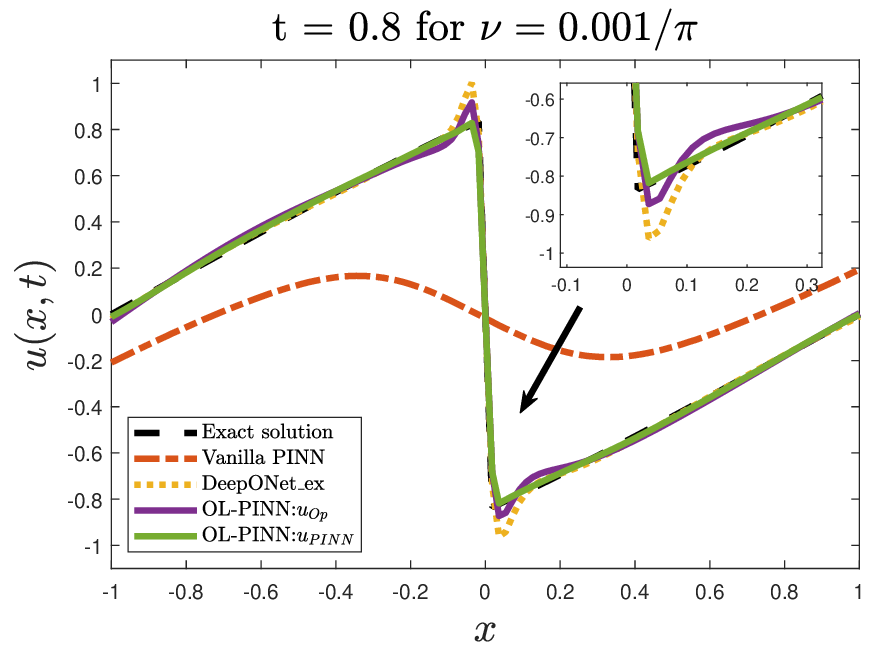}}
		\subfloat[\label{fig5:d}$\nu = 0$ without BCs]{
			\includegraphics[scale=0.45]{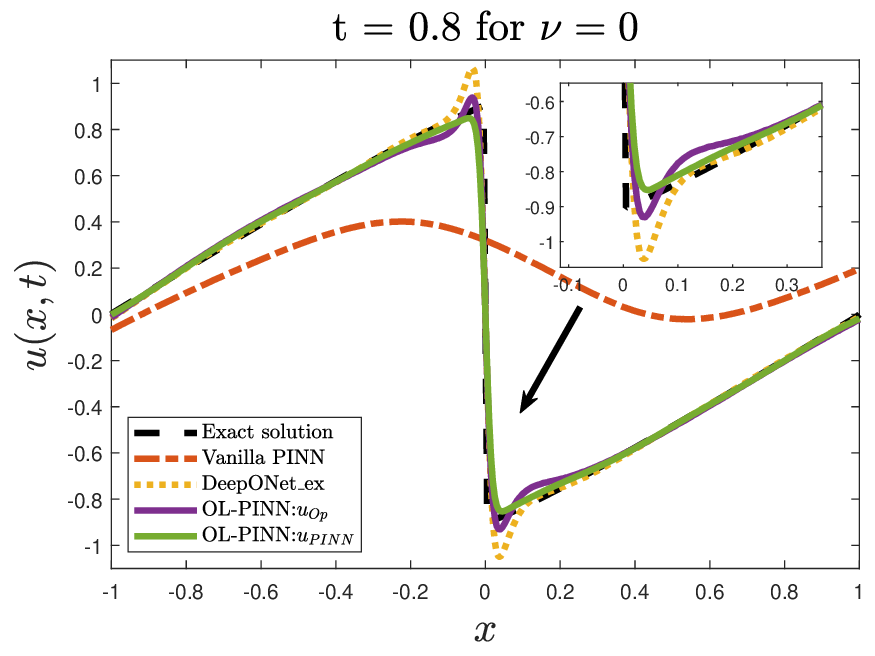}}
		\caption{Example 2, Case I: Comparison of the prediction at time $t=0.8$ for different models. Here we use $51\times 10$ uniformly distributed residual points for the $(x,t)$ domain. (a) Predictions for $\nu = 0.001/\pi$ with BCs. (b) Predictions for $\nu = 0$ with BCs. (c) Predictions for $\nu = 0.001/\pi$ without BC. (d) Predictions for $\nu = 0$ without BC.}
		\label{nu_ex}
\end{figure}

We train the OL-PINN by using Adam optimizer  with 20000 epochs, learning rate 0.001 and \emph{\bm{$51\times 10$} uniformly distributed residual points} for the $(x,t)$ domain.
The results at time $t=0.8$ for both the viscid and inviscid cases are shown in left and right plots of Fig. \ref{nu_ex}, respectively. 
Observe that the vanilla PINN fails to converge to the reference solutions. We point out here that even the vanilla PINN is trained with a much larger number of epoch, it still fails to converge. However, this is resolved by using the present method, and promisingly, we see that \emph{the sharpness can be well captured without oscillations} for the PINN prediction of OL-PINN.

\begin{figure}[htbp]
	\centering
	\subfloat{
		\includegraphics[scale=0.45]{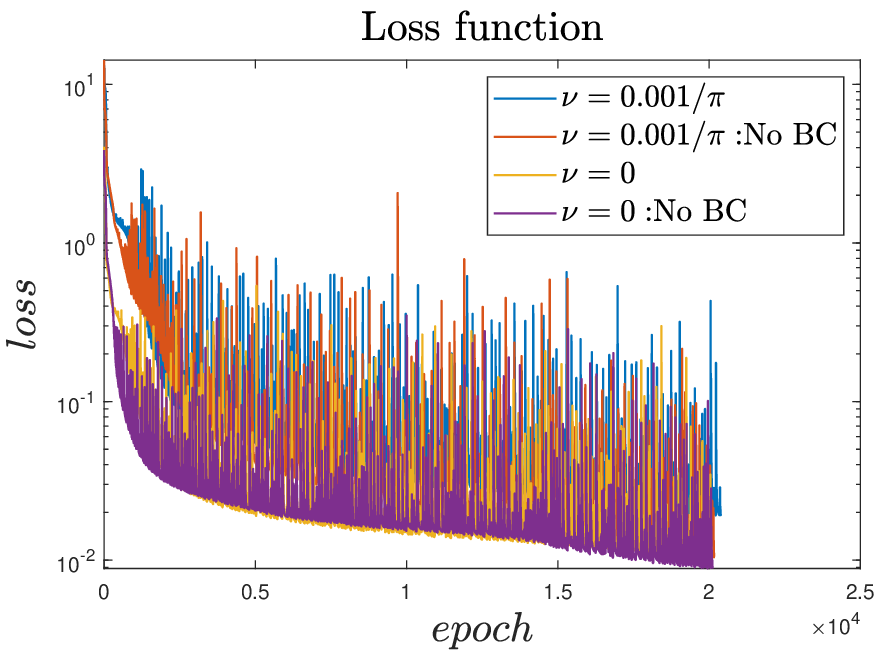}}
  \subfloat{
		\includegraphics[scale=0.45]{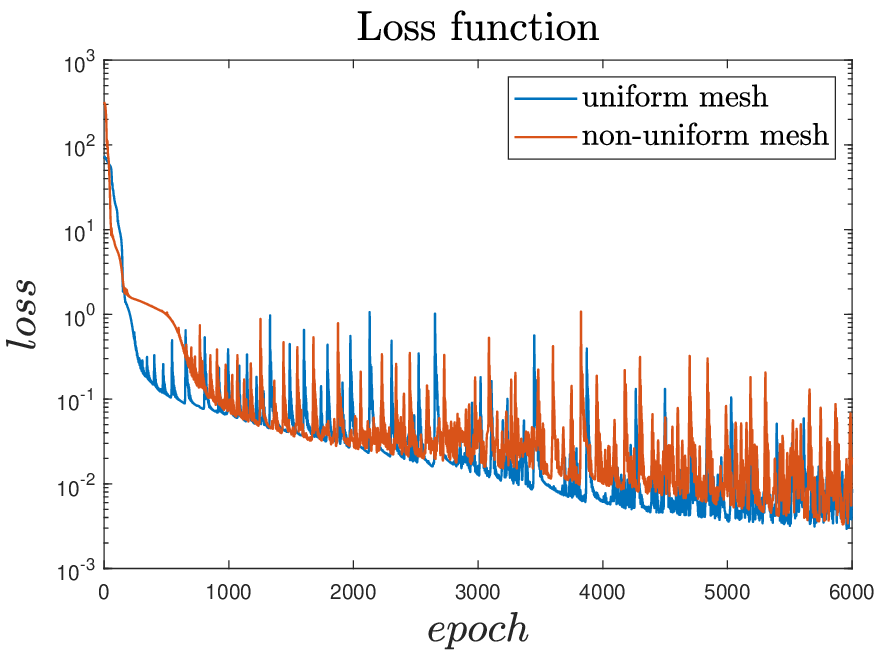}}
	\caption{Example 2: Loss vs number of epochs. Left: case I, right: case II.}
	\label{lossvsepoch:burgers}
\end{figure}

Using almost the same setup, we further consider the \emph{ill-posed} problem, namely, we consider the Burgers equation without using the boundary conditions. This kind of problem cannot be solved even by using the classical method. The loss histories in this case as well as the previous case are presented in the left plot of Fig. \ref{lossvsepoch:burgers}.
As we did previously, we present the profiles at time $t=0.8$ in Figs. \ref{fig5:c} and \ref{fig5:d} showing again that good predictions are obtained by using our method.

To quantitatively illustrate the effectiveness of the present method, we also compare the relative $L^2$-error for different models shown in Table \ref{tab_nu}. We observe that both the predictions of $u_{PINN}$ and $u_{Op}$ using OL-PINN yield good accuracy. Moreover, the small standard deviations of $u_{PINN}$ and $u_{Op}$ indicate that the training process of the present method is stable. This means that the OL-PINN framework not only improves the accuracy but also stabilizes the training process.

\begin{table}[htbp]
		\centering
		\begin{tabular}{|c|c|c|c|c|}
			\hline
    			{ Model} & Vanilla PINN  &DeepONet\_ex& OL-PINN:$u_{PINN}$  &OL-PINN:$u_{Op}$  \\
			\hline  { $\nu = 0.001/\pi$ } & $82.19 \pm 30.16 $ & $6.73\pm7.65 $& $\bm{3.49 \pm 0.67 }$ & $4.47  \pm 0.60$ \\
			\hline  { $\nu = 0.001/\pi$, no BCs} & $100.2  \pm 76.16$ & $ 6.73\pm7.65  $& $\bm{3.86 \pm 1.01} $ & $4.45  \pm 0.50$ \\
			\hline  { $\nu=0 $ } & $69.78 \pm 13.70 $ & $8.26\pm2.80$& $7.87 \pm 0.70 $ & $\bm{7.54  \pm 0.66} $ \\
			\hline  { $\nu = 0$, no BCs} & $72.55 \pm 28.54 $ & $8.26\pm2.80$& $7.91 \pm 0.88$ & $\bm{7.34  \pm 0.52}$ \\
			\hline
		\end{tabular}\caption{Example 2, Case I: Mean and standard deviation of the relative $L^2(\%)$ test error for each model.}
		\label{tab_nu}
	\end{table}

\subsubsection{Case II: Prediction for unseen $t$}\label{section:caseII}
In real applications, such as weather forecast and oceanographic monitoring and forecasting, we have lots of history data and want to make predictions for a short-term future. Motivated by this, we would like to train a DeepONet by using the history data, and then make predictions using data and/or physics. This is exactly one of the application of OL-PINN.

\begin{figure}[htbp]
		\centering
		\subfloat[\label{fig:b} Uniform $\{x^r\}$]{
			\includegraphics[scale=0.45]{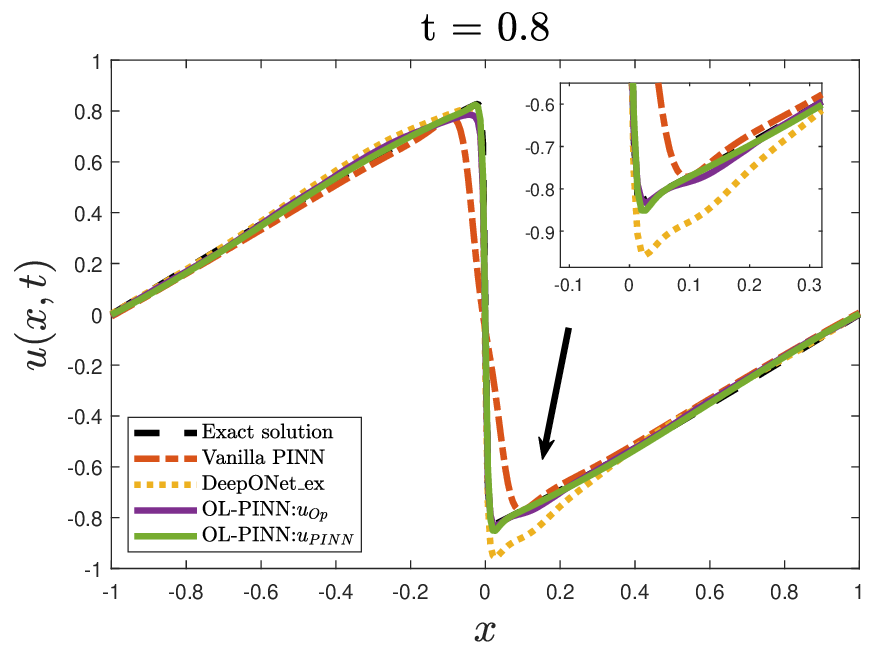}}
		\subfloat[\label{fig:d} clustered $\{x^r\}$]{
			\includegraphics[scale=0.45]{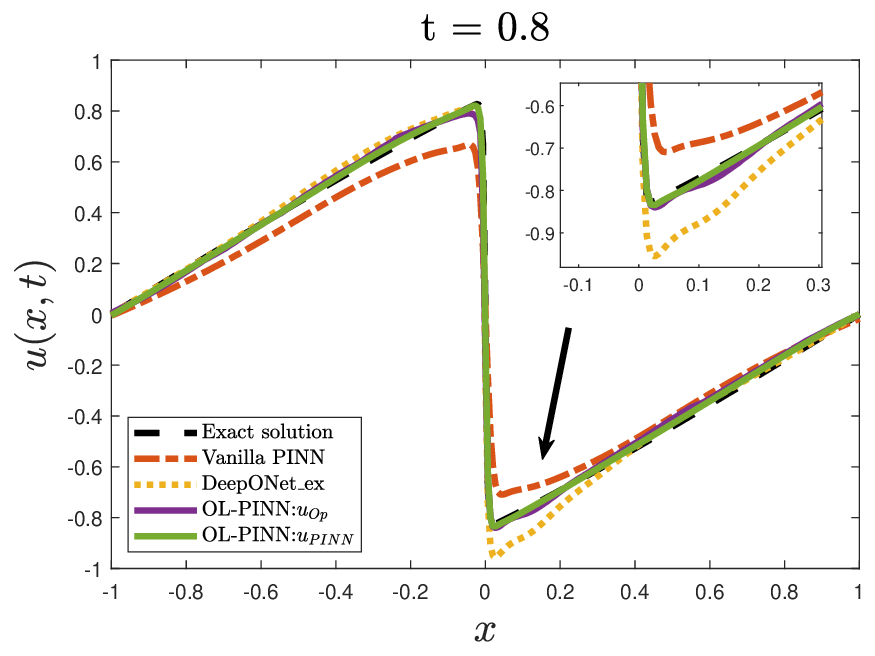}}
		\caption{Example 2, Case II: Comparison of the prediction at time $t=0.8$ for different models. Here we use $21\times 3$ residual points for the $(x,t)$ domain. (a) The residual points are uniformly distributed. (b) The residual points are non-uniformly  distributed.}
		\label{t_ex}
	\end{figure}

Therefore, in this subsection we again use the Burgers equation as the benchmark. 
Note that for a smooth initial condition, the solution of the Burgers equation becomes sharper as $t$ grows if the viscosity $\nu$ is tiny.
Now we use the data of $t\in[0, 0.6]$ to train the DeepONet, and then solve the Burgers equation for $t \in [0.6,0.8]$. In this case, we \emph{only use \bm{$21\times 3$} residual points} for the $(x,t)$ domain.
We test two different distributed residual points, i.e., uniformly and non-uniformly distributed residual points. For the non-uniformly residual points, we use 11 uniformly distributed points for $x \in [-0.2,0.2]$ while the remaining points are uniformly distributed in the rest of interval.
We use the Adam optimizer with 6000 epochs and learning rate 0.001 for the training. The training losses are shown in the right plot of Fig. \ref{lossvsepoch:burgers} while the result at time $t=0.8$ is shown in Fig. \ref{t_ex}. 
The mean and standard deviation of the relative L2 error for these two cases are presented in Table \ref{result_t}. Again, we see from these results that the prediction of $u_{PINN}$ and $u_{Op}$ are much better than the ones of the vanilla PINN and the extrapolation of DeepONet showing that the OL-PINN is a high accuracy, efficient and stable approach for sharp problems.

\begin{table}[htbp]
		\centering
		\begin{tabular}{|c|c|c|c|c|}
			\hline
            Model & Vanilla  PINN  & DeepONet\_{ex} & OL-PINN: $u_{PINN}$ & OL-PINN: $u_{Op}$ \\
			\hline  {uniform mesh } & $ 21.92 \pm 2.43 $ & $7.42  \pm 0.14  $& $3.22 \pm 0.02 $ & $\bm{2.68 \pm 0.04} $\\
			\hline  {non-uniform mesh } & $20.24 \pm 11.39 $ & $7.42  \pm 0.14$ & $3.32  \pm 0.70$& $\bm{2.68 \pm 0.37}$\\

			\hline
		\end{tabular}\caption{Example 2, Case II: Mean and standard deviation of the relative $L^2(\%)$ test error for each model.}
		\label{result_t}
	\end{table}

\subsection{Navier-Stokes equations}
We consider in this section the Navier-Stokes equations. We solve the steady state of the lid-driven cavity flow in Subsection \ref{subsection:cavity} while we solve the time-dependent two-dimensional Navier-Stokes equation in vorticity form in \ref{subsection:vorticity}.

\subsubsection{Example 3:  Lid-driven cavity flow}\label{subsection:cavity}
In this section we consider the steady state of the 2D incompressible cavity flow in a square cavity (i.e., $x, y \in [0, 1]^2$), which is a classical CFD problem and is described by the Navier-Stokes equations
%
\begin{equation}\label{eq:stokes}
    \left\{ 
    \begin{aligned}
        & \boldsymbol{u}(x,y) \cdot \nabla \boldsymbol{u}(x,y)  + \nabla p(x,y) - \nu \Delta \boldsymbol{u}(x,y) = 0, \\
        &\nabla \cdot \boldsymbol{u}(x,y) =0, 
    \end{aligned}
    \right. \quad 
    (x,y) \in \Omega:=(0,1)^2
\end{equation}
 with boundary conditions 
 \begin{equation}\label{eq:NS-bcs}
     	\boldsymbol{u}|_{y=1} =1 - \frac{\cosh(10(x-0.5))}{\cosh{(5)}},\; \boldsymbol{u}|_{\partial \Omega \setminus y=1}=0,\; \boldsymbol{v}|_{\partial \Omega}=0.
 \end{equation}
where $\boldsymbol{u}=(u,v)$ is the velocity field, $\nu$ is the kinematic viscosity, $p(x,y)$ is the pressure. $Re$ is the the Reynolds number and $\nu = \frac{1}{Re}$. 
%
Note that the equation \eqref{eq:stokes} with BCs \eqref{eq:NS-bcs} is not a well-posed problem since it has multiple solutions~\cite{wang2023solution}. It is unstable if using the vanilla PINN or a plain classical numerical method.

In this example, we employ the OL-PINN to solve the above equation with $Re=1000$, and pretrain the DeepONet for the $Re \in \{400,420,440,...,680\}$. The data is obtained by using the spectral element method~\cite{karniadakis2005spectral}. For the divergence free condition, i.e., $\nabla \cdot \boldsymbol{u} = 0$, which is the continuity equation for incompressible fluids describing the conservation of mass of the fluid, we use two different approaches:
\begin{itemize}
    \item Case I: The first approach used here is to directly treat the condition $\nabla \cdot \boldsymbol{u} = 0$ as an additional equation, which is served as one loss term. In this case, the divergence free condition is satisfied in the collocation sense.
    \item Case II: For the second one, we adopt the technique given in \cite{raissi2019physics}. Namely, we introduce an auxiliary function $\psi(x,y)$ as the output of the PINN and set 
    \begin{align*}
		u=\psi_y, \quad v=-\psi_x.
	\end{align*}
 Then, one can easily checked that the condition $\nabla \cdot \boldsymbol{u} = 0$ is satisfied automatically.
\end{itemize}
We comment here the advantage and disadvantage of the above two approaches. For the first one, the divergence free condition is not exactly satisfied and only imposed in a set of discrete collocation points, which would make the training to be more difficult. While for the second approach, the divergence free condition is automatically satisfied, and this would accelerate the training process. However, in the training of neural networks, the computation of derivatives is time consuming, leading to a less computationally efficient in each epoch by using the second approach since it requires computing higher order derivatives with respect to the input for the second approach.

To this end, we use \emph{about \bm{$10000$} residual points} and train the networks with 200000 epochs for Case I while we use \emph{only about \bm{$900$} residual points} and train the networks with 80000 epochs for Case II. In both cases we use the Adam optimizer with learning rate 0.0005.
Note that, in the work proposed in \cite{he2023artificial} where the artificial viscosity is used, 40000 residual points and 600000 epochs are used for the problem with $Re = 600$. We see that less than $1/40$ residual points are used in our case II.

\begin{figure}[htbp]
		\centering
		\subfloat[Reference solution\label{fig7:a}]{
			\includegraphics[scale=0.44, trim=1.5cm  0  1.5cm 1cm]{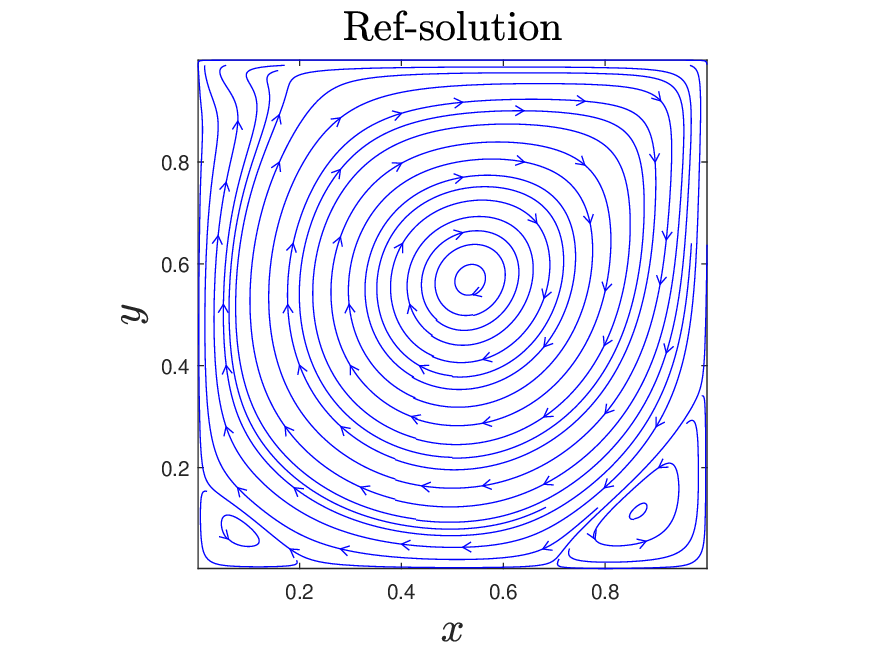}}
		\subfloat[Vanilla PINN solution\label{fig7:b}]{
			\includegraphics[scale=0.44, trim=1.5cm  0  1.5cm 1cm]{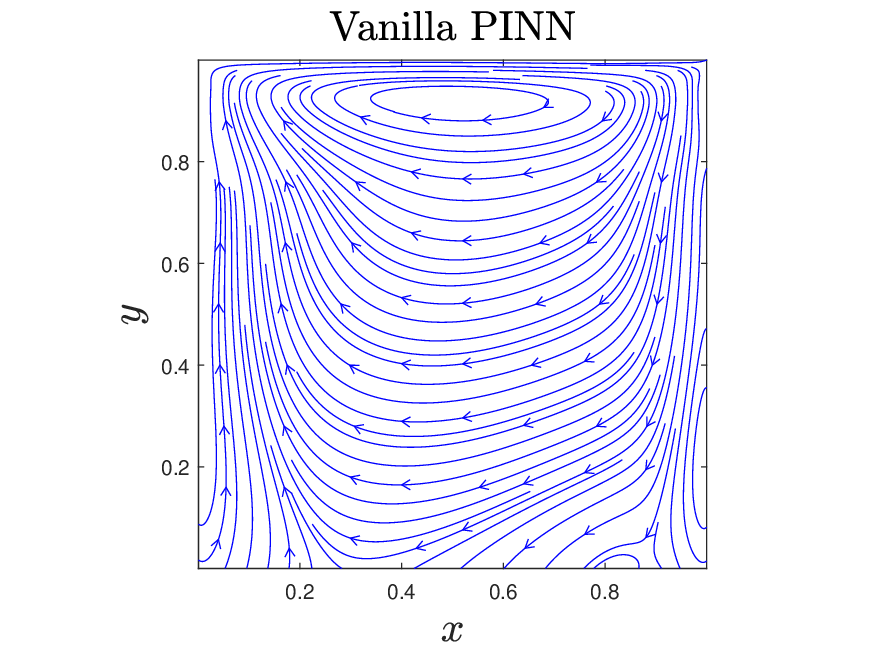}}
		\subfloat[DeepONet extrapolation\label{fig7:c}]{
			\includegraphics[scale=0.44, trim=1.5cm  0  1.5cm 1cm]{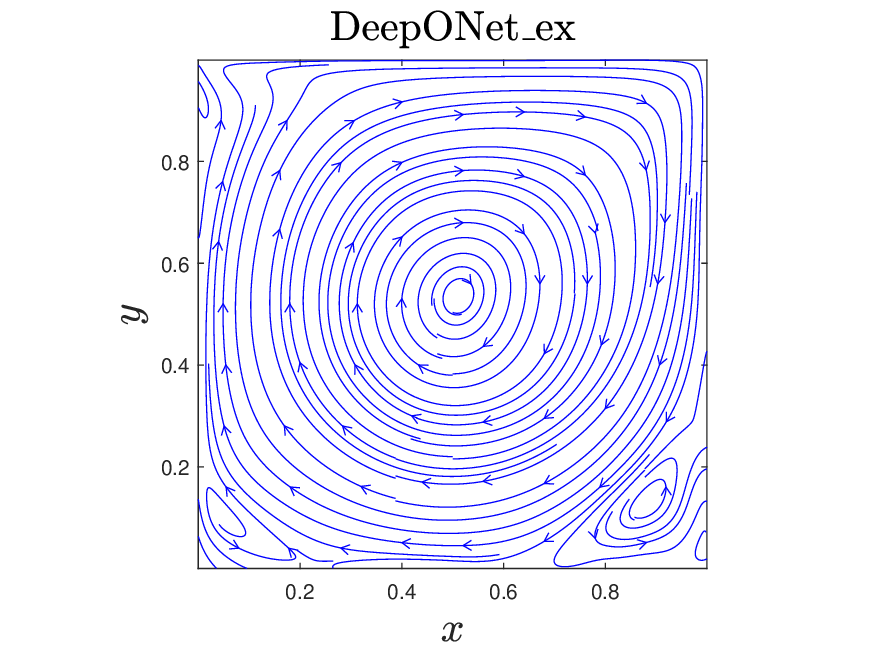}}
		
		\quad
		\subfloat[OL-PINN: $u_{PINN}$\label{fig7:d}]{
			\includegraphics[scale=0.44, trim=1.5cm  0  1.5cm 0]{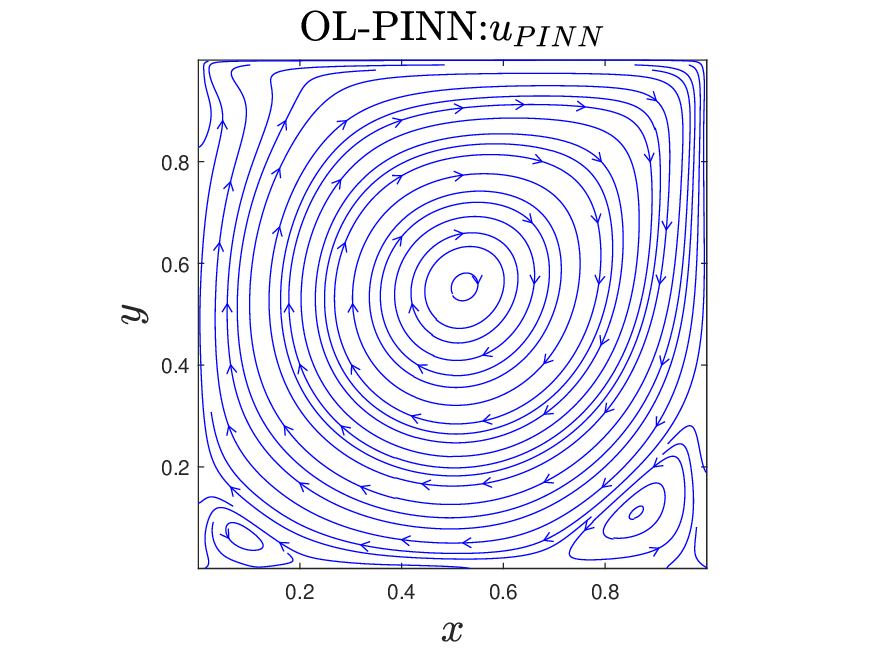}}    
		\subfloat[OL-PINN: $u_{OP}$\label{fig7:e}]{
			\includegraphics[scale=0.44, trim=1.5cm  0  1.5cm 0]{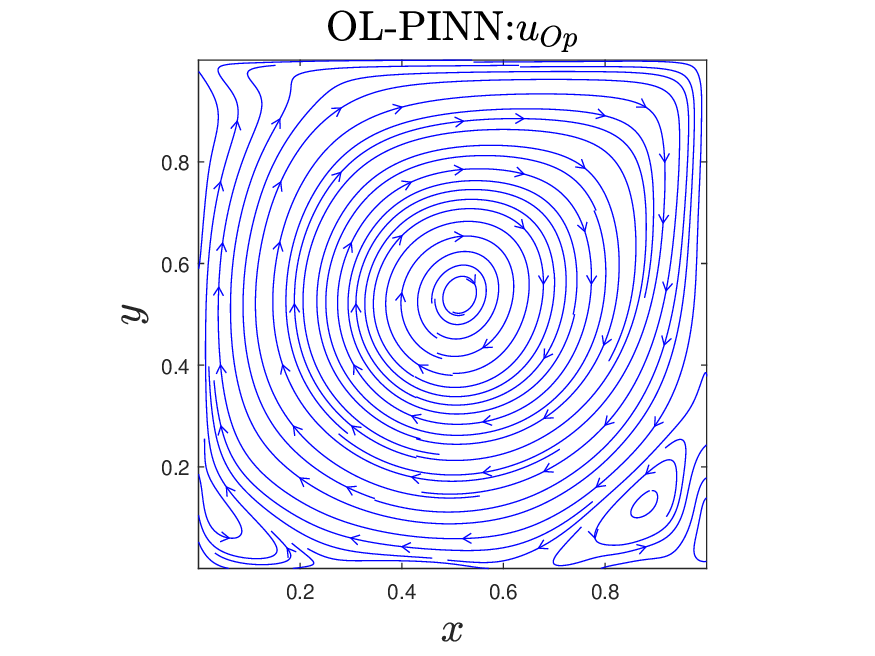}}
		\subfloat[Loss vs number of epochs\label{fig7:f}]{
			\includegraphics[scale=0.44, trim=1.5cm  0  1.5cm 0]{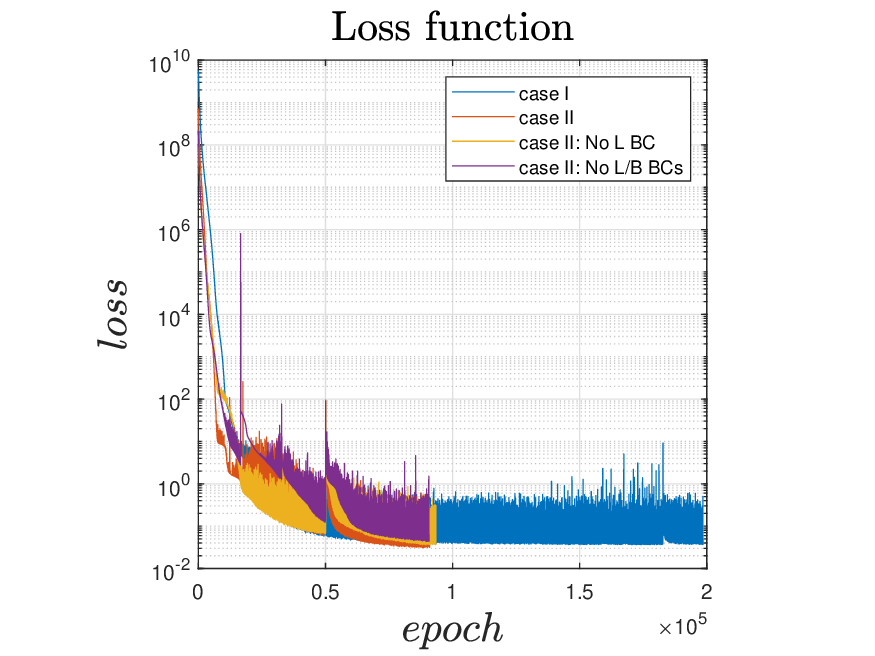}}
   
		\caption{Example 3, Case I: (a)-(e): Comparison of the prediction of streamlines for different models with $Re = 1000$. Here we use $101\times 101$ residual points for the $(x,y)$ domain. The divergence free condition is employed as a soft constrain. (f): Loss vs number of epochs.}
		\label{cavity_ex1}
	\end{figure}

Now let us first focus on Case I. In this case, we use uniformly distributed residual points and the Adam optimizer with  and learning rate 0.0005, and plot the streamlines for the velocity for different models in Figs. \ref{fig7:a}-\ref{fig7:e}. The loss history is given in Fig. \ref{fig7:f}, and the mean and standard deviation of the relative error are presented in Table \ref{table_cavity}. Again, we observe that the vanilla PINN fails to solve the proposed problem (The relative error is larger than 80\%). However, we obtain good predictions $u_{PINN}$ and $u_{Op}$ by using our method, and the prediction of $u_{PINN}$ has the best performance among all models. It is hard to tell from Fig. \ref{fig7:c} and Fig. \ref{fig7:e} that which one is better, $u_{Op}$ or the extrapolation of the DeepONet? But we see from Table \ref{table_cavity} the prediction of $u_{Op}$ is more accurate, and the training process is very stable using the present method.

\begin{figure}[htbp]
		\centering
		\subfloat[Uniform $\{x^r\}$\label{fig9:a}]{
			\includegraphics[scale=0.44 ,trim=1.5cm  0  1.5cm 1cm]{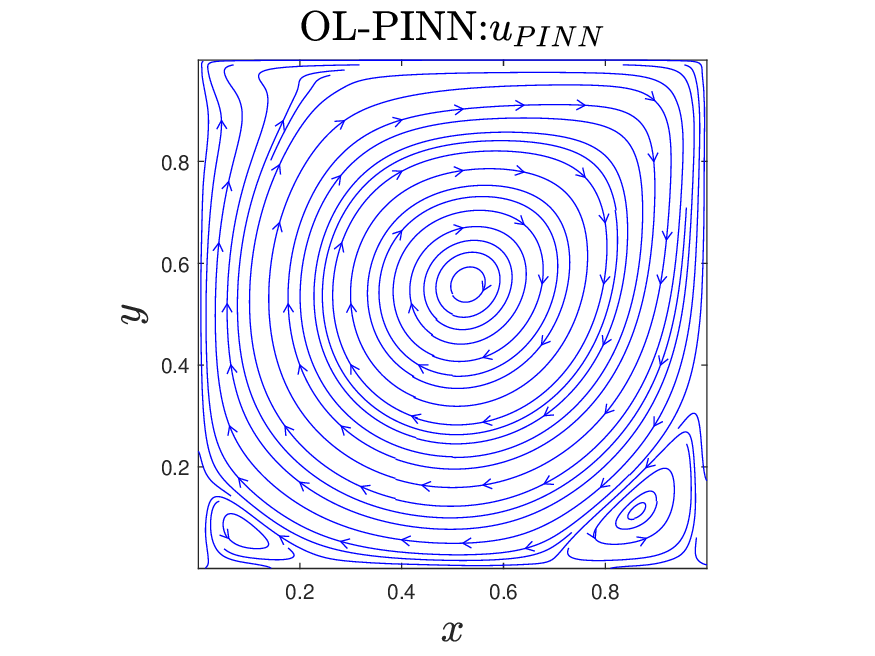}}
		\subfloat[Uniform $\{x^r\}$, no L BC\label{fig9:b}]{
			\includegraphics[scale=0.44,trim=1.5cm  0  1.5cm 1cm]{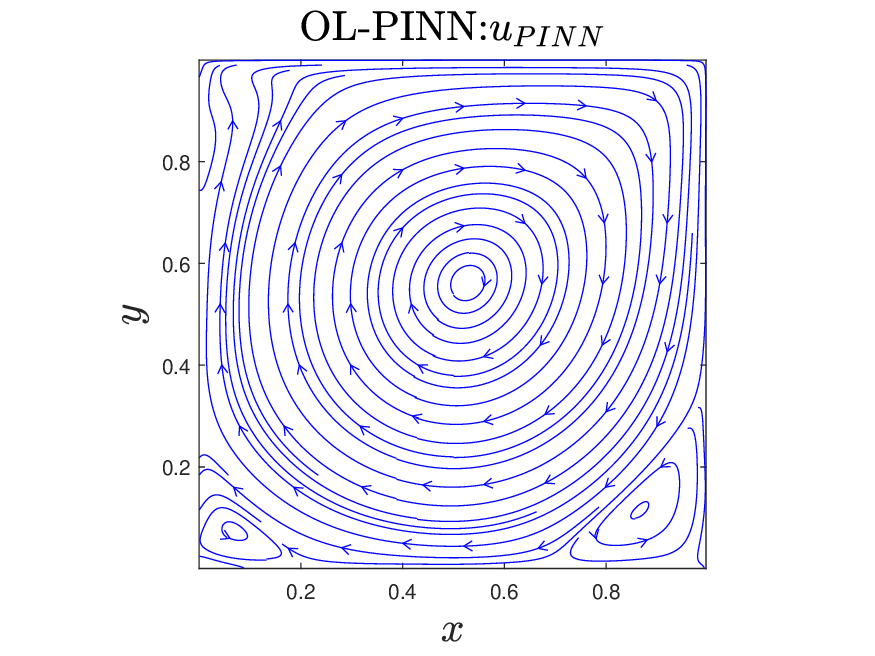}}
		\subfloat[Uniform $\{x^r\}$, no L/B BCs\label{fig9:c}]{
			\includegraphics[scale=0.44,trim=1.5cm  0  1.5cm 1cm]{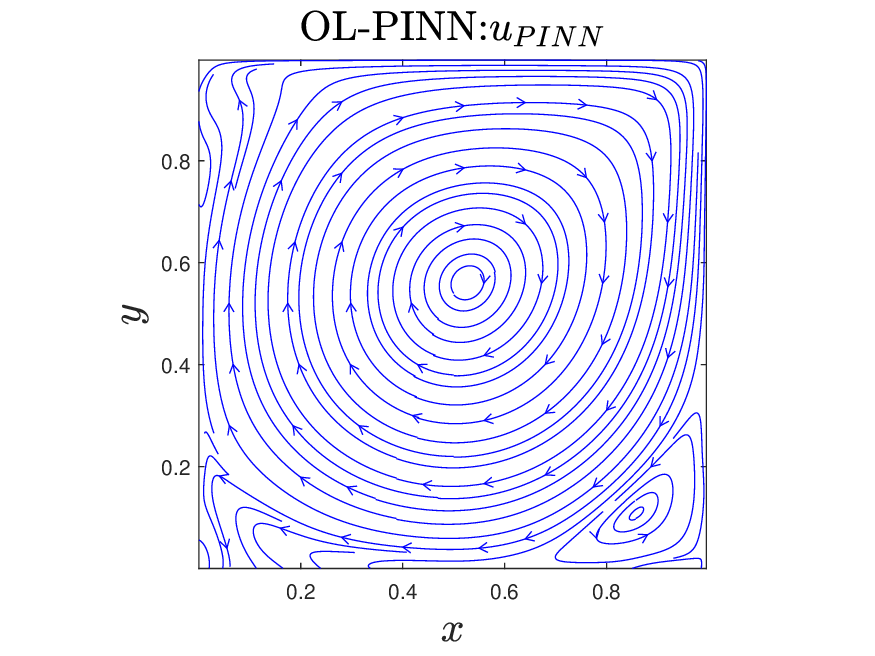}}
		\quad
		\subfloat[Random $\{x^r\}$\label{fig9:d}]{
			\includegraphics[scale=0.44,trim=1.5cm  0  1.5cm 0]{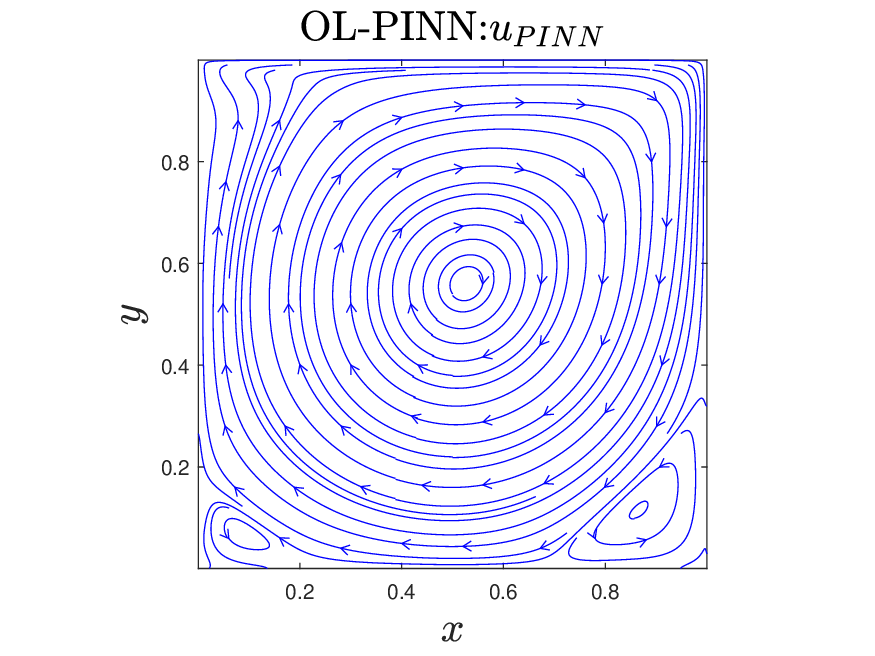}}
		\subfloat[Random $\{x^r\}$, no L BC\label{fig9:e}]{
			\includegraphics[scale=0.44,trim=1.5cm  0  1.5cm 0]{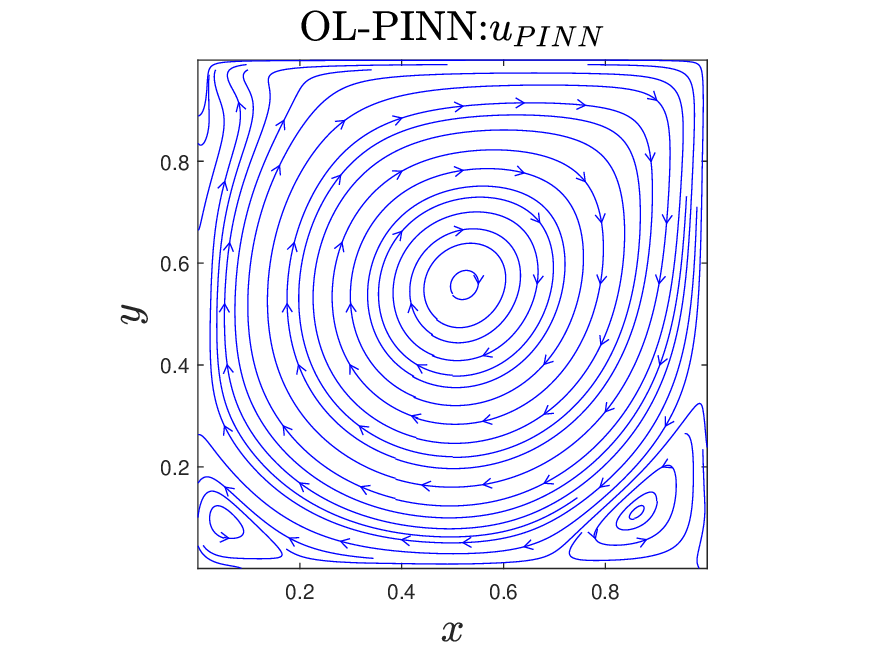}}
		\subfloat[Random $\{x^r\}$, no L/B BCs\label{fig9:f}]{
			\includegraphics[scale=0.44,trim=1.5cm  0  1.5cm 0]{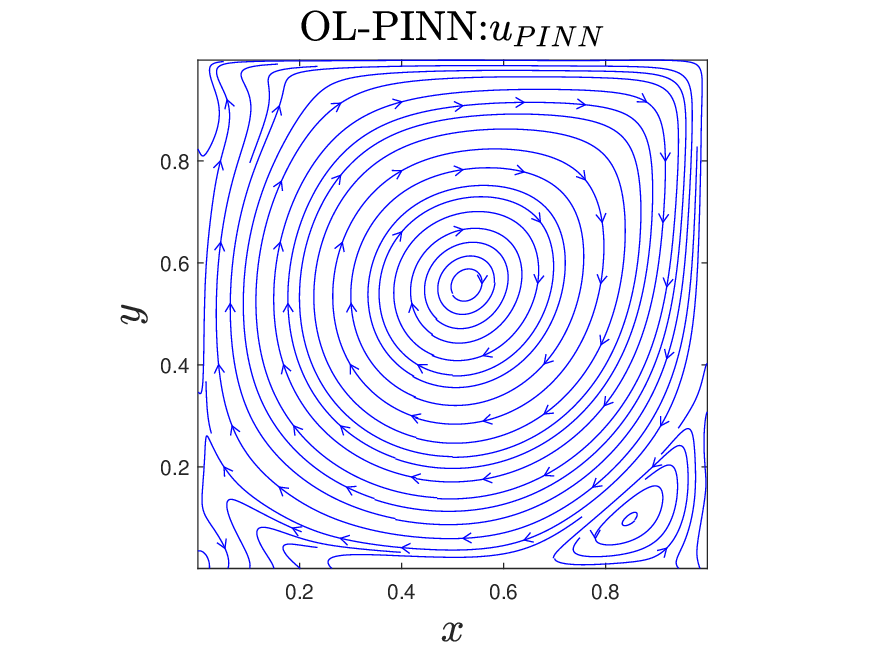}}
		\caption{Example 3, Case II: Prediction of streamlines for the $u_{PINN}$ model with $Re = 1000$. Here we use $31\times 31$ residual points for the $(x,y)$ domain. The divergence free condition is severed as a hard constrain.}
		\label{cavity_ex2}
	\end{figure}

We now turn to Case II. In this case, we test both uniformly and randomly distributed residual points, and the corresponding results of the streamlines of $u_{PINN}$ are shown in Fig. \ref{fig9:a} and \ref{fig9:d}, respectively. 
Furthermore, as done for the Burgers equation, we also consider the equation \eqref{eq:stokes} with insufficient BCs, for instance, here we consider two cases: (i) no left BC (no L BC), (ii) no left and bottom BCs (no L/B BCs). 
We present the results obtained by using uniform and clustered residual points in in Figs. \ref{fig9:b}-\ref{fig9:c} and Figs. \ref{fig9:e}-\ref{fig9:f}, respectively. Similarly, the result of the relative $L^2$ errors are given in Table \ref{table_cavity} and \ref{table_cavity2}.
For all these tests, the results indicate that we can use the present method to resolve the (ill-posed) equation \eqref{eq:stokes} with high Reynolds number ($Re = 1000$) using a small number of residual points (ex., about 900 residual points). Also, we see from Table \ref{table_cavity} and \ref{table_cavity2} that the prediction of $u_{PINN}$ has supreme accuracy among all models. The relative error is reduced to less than 5\% for the PINN solutions. In addition, we can resolve the problem even with only partial BCs. This property shows the potential of the present method in solving realistic complex problems.
Moreover, we see that \emph{the vortices are well captured} except the cases without L/B BCs.



 \begin{table}[htbp]
    \centering
    \begin{tabular}{|c|c|c|c|c|c|}
 \hline \multicolumn{2}{|c|}{Model } & Vanilla PINN& DeepONet\_ex& OL-PINN:$u_{PINN}$& OL-PINN:$u_{Op}$ \\
\hline \multirow{2}{*}{\makecell{Case I:\\ Uniform $\{x^r\}$}} & u & $83.11\pm1.84$ & $13.64\pm14.49$ & \bm{$5.82 \pm 0.04 $} & $12.08\pm0.01$ \\
\cline { 2 - 6 } & v & $93.11\pm4.86$ & $12.95\pm8.67$ & \bm{$4.45 \pm 0.04$} & $9.23\pm0.01$ \\

\hline \multirow{2}{*}{\makecell{Case II:\\ Uniform $\{x^r\}$}} & u & $75.22\pm18.95$ & $13.64\pm14.49$ & \bm{$4.38 \pm 2.23 $} & $12.13\pm0.03$ \\
\cline { 2 - 6 } & v & $80.27\pm18.29$ &$12.09\pm0.01$ & \bm{$3.85 \pm 1.45$} & $9.21\pm0.04$ \\

\hline \multirow{2}{*}{\makecell{Case II:\\ Random $\{x^r\}$}} & u & $94.95\pm17.41$ & $13.64\pm14.49$ & \bm{$3.82\pm0.19$} & $12.09\pm0.02$ \\
\cline { 2 - 6 } & v & $98.00\pm7.53$ & $12.95\pm8.67$ & \bm{$3.47 \pm 0.45$} & $9.22\pm0.03$ \\

\hline
\end{tabular}
    \caption{Example 3: Mean and standard deviation of the relative $L^2(\%)$ test error for each model with \emph{all BCs} \eqref{eq:NS-bcs}.}
    \label{table_cavity}
\end{table}

\begin{table}[htbp]

    \centering
    \begin{tabular}{|c|c|c|c|c|c|}
 \hline \multicolumn{2}{|c|}{Model } & Vanilla PINN& DeepONet\_ex& OL-PINN:$u_{PINN}$& OL-PINN:$u_{Op}$ \\

\hline \multirow{2}{*} {\makecell{ Case II: No L BC \\ Uniform $\{x^r\}$}} & u & $53.50\pm23.20$ & $13.64\pm14.49$ & \bm{$3.69\pm0.13$} & $12.09\pm0.01$ \\
\cline { 2 - 6 } & v & $48.62\pm15.81$ & $12.95\pm8.67$ & \bm{$3.21\pm0.05$} & $9.35\pm0.14$ \\

\hline \multirow{2}{*} {\makecell{Case II:No L BC \\ Random $\{x^r\}$}} & u & $58.45\pm32.93$ & $13.64\pm14.49$ & \bm{$4.29\pm0.09$} & $12.11\pm0.03$ \\
\cline { 2 - 6 } & v & $48.03\pm39.49$ & $12.95\pm8.67$ & \bm{$3.60\pm0.30$} & $9.33\pm0.08$ \\

\hline \multirow{2}{*} {\makecell{Case II: No L/B BCs \\  Uniform $\{x^r\}$}} & u & $82.32\pm3.63$ & $13.64\pm14.49$ & \bm{$3.97\pm0.20$ }& $12.10\pm0.10$ \\
\cline { 2 - 6 } & v & $79.14\pm3.69$ & $12.95\pm8.67$ & \bm{$3.89\pm0.20$} & $9.66\pm0.08$ \\

\hline \multirow{2}{*} {\makecell{Case II:No L/B BCs \\ Random $\{x^r\}$}} & u & $89.44\pm30.43$ & $13.64\pm14.49$ & \bm{$4.40\pm0.19$} & $12.10\pm0.11$ \\
\cline { 2 - 6 } & v & $79.43\pm28.60$ & $12.95\pm8.67$ & \bm{$4.20\pm0.21$ }& $9.59\pm0.10$ \\

\hline
\end{tabular}
    \caption{Example 3: Mean and standard deviation of the relative $L^2(\%)$ test error for each model with \emph{partial BCs}.}
    \label{table_cavity2}
\end{table}

\subsubsection{Exampe 4: Navier-Stokes equation in vorticity form}\label{subsection:vorticity}
Following the problem setting in \cite{li2020fourier}, we consider in this subsection the time-dependent two-dimensional Navier-Stokes equation for a viscous, incompressible fluid in vorticity form on the unit torus:
\begin{equation}
 \left\{
		\begin{aligned}
		 &\partial_t w(x,y, t)+\boldsymbol{u}(x,y, t) \cdot \nabla w(x,y, t) =\nu \Delta w(x,y,t)+f(x,y),  & & x,y \in(0,1)^2, t \in(0, T], \\
		 &\nabla \cdot \boldsymbol{u}(x,y, t) =0,  & & x,y \in(0,1)^2, t \in[0, T] \label{div_ns},\\
		 & w(x,y, 0) =w_0(x,y), & & x,y \in(0,1)^2,
		\end{aligned}
  \right.
\end{equation}
where $\nu = 0.001$, $\boldsymbol{u} \in C\left([0, T] ; H_{\text {per }}^1\left((0,1)^2 ; \mathbb{R}^2\right)\right)$ is the velocity field, $w=\nabla \times \boldsymbol{u}$ is the vorticity, $w_0 \in L_{\text {per }}^2\left((0,1)^2 ; \mathbb{R}\right)$ is the initial vorticity,  and the forcing function is given by $f(x)=0.1 \sin (2 \pi(x+y))+0.1 \cos (2 \pi(x+y)) $, $f \in$ $L_{\text {per }}^2\left((0,1)^2; \mathbb{R}\right)$. 

In this example, we pretrain a DeepONet to learn the operator  $w|_{(0,1)^2 \times(5,6]} \mapsto w|_{(0,1)^2 \times(6, 7]}$ with the initial condition $w_0(x)$ generated by a Gaussian random field $\mathcal{N}\left(0,~8^{4}(-\Delta+64 I)^{-5}\right)$, and solve the problem with the initial condition function $w_0(x)$ generated by a Gaussian random field $\mathcal{N}\left(0,~4^{1/5}(-\Delta+16 I)^{-1.2}\right)$.
Here we use the second approach for the condition $\nabla \cdot \boldsymbol{u} = 0$ used in the previous subsection, i.e., we enforce the divergence free condition in the neural network function. For the residual points, we use the spatial-temporal resolution to be $22\times 22 \times 20$.
Additionally, we use 101 equally distributed points for $u$ at each boundary and set the spatial resolution to be $64 \times 64$ and $190 \times 190$ for the initial condition for PINN and NN, respectively. We train the networks using the Adam optimizer with 50000 epochs learning rate 0.001.
Similarly as considered for previous examples, we consider the following two cases:
\begin{itemize}
    \item Case I: well-posed problem with sufficient BCs.
    \item Case II: ill-posed problem with insufficient BCs. In this case, we use the boundary conditions that only $w$ is periodic.
\end{itemize}
 
\begin{figure}[htbp]
		\centering
  \subfloat[Ref. solution\label{fig8:a}]{
			\includegraphics[scale=0.25]{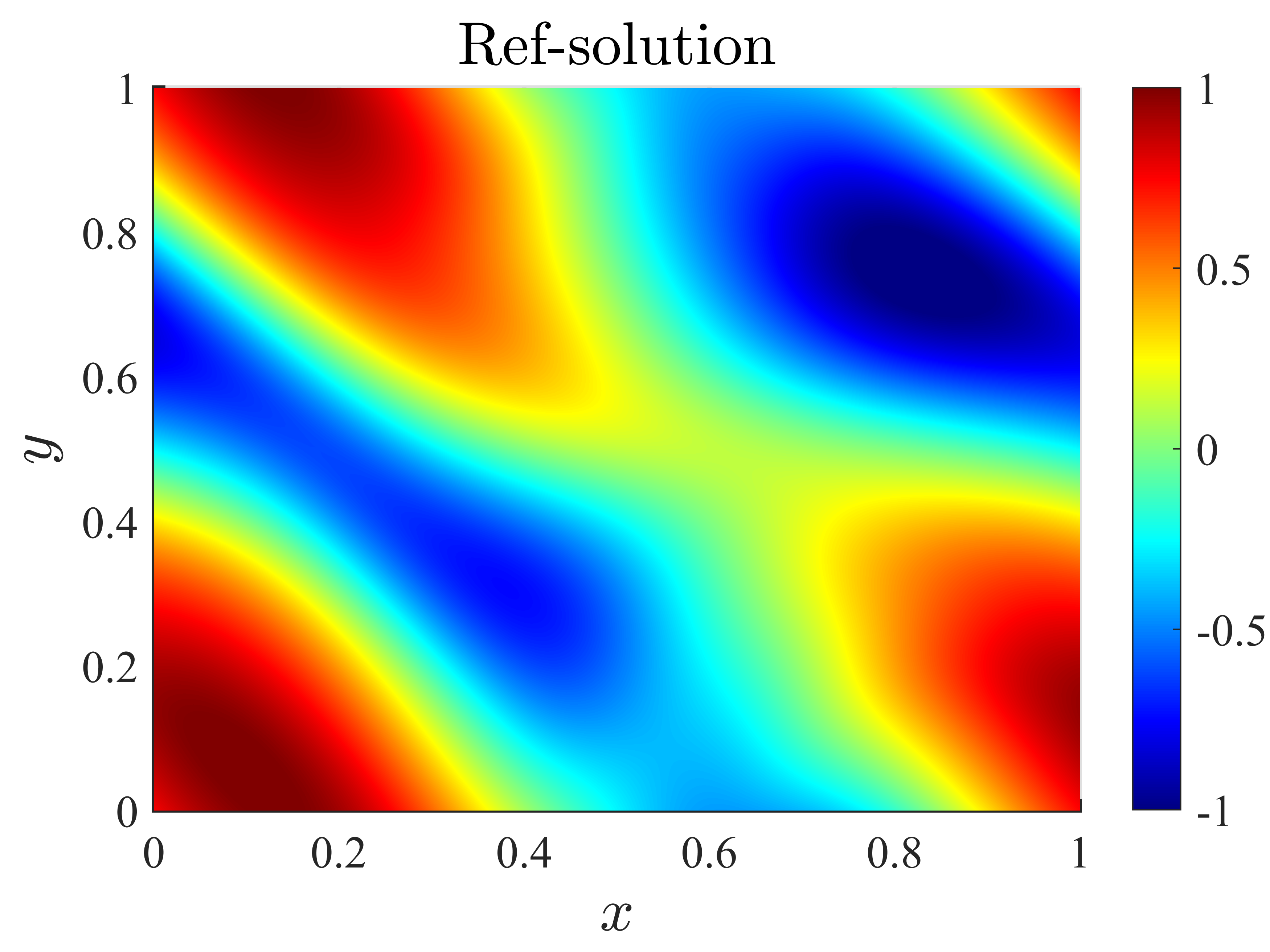}}
   \subfloat[$|u_{ref} - u_{PINN}|$\label{fig8:b}]{
			
			\includegraphics[scale=0.25]{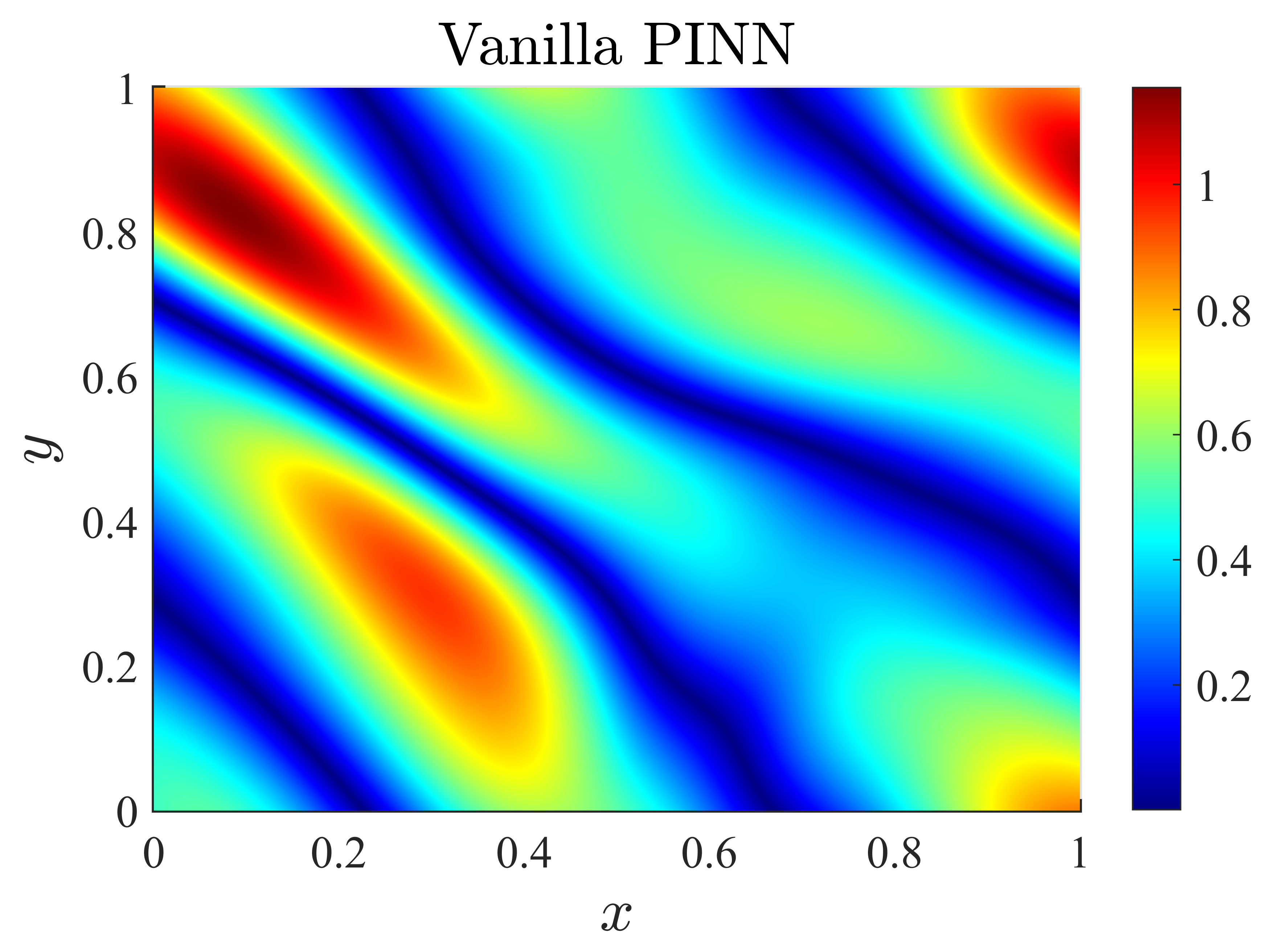}}
		\subfloat[$|u_{ref} - u_{ex}|$\label{fig8:c}]{
			
			\includegraphics[scale=0.25]{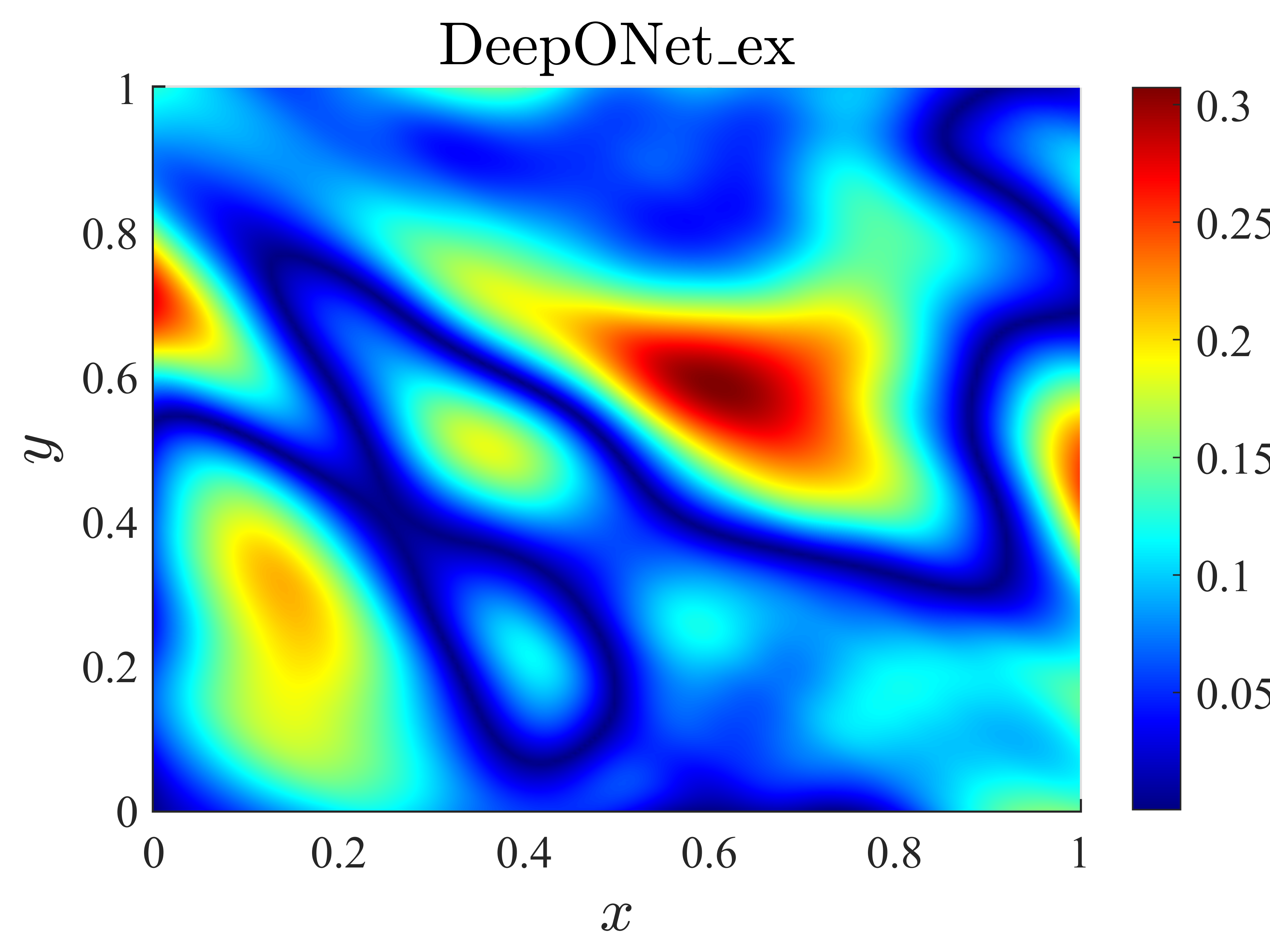}}
		\subfloat[$|u_{ref} - u_{PINN}|$ \label{fig8:d}]{
			
			\includegraphics[scale=0.25]{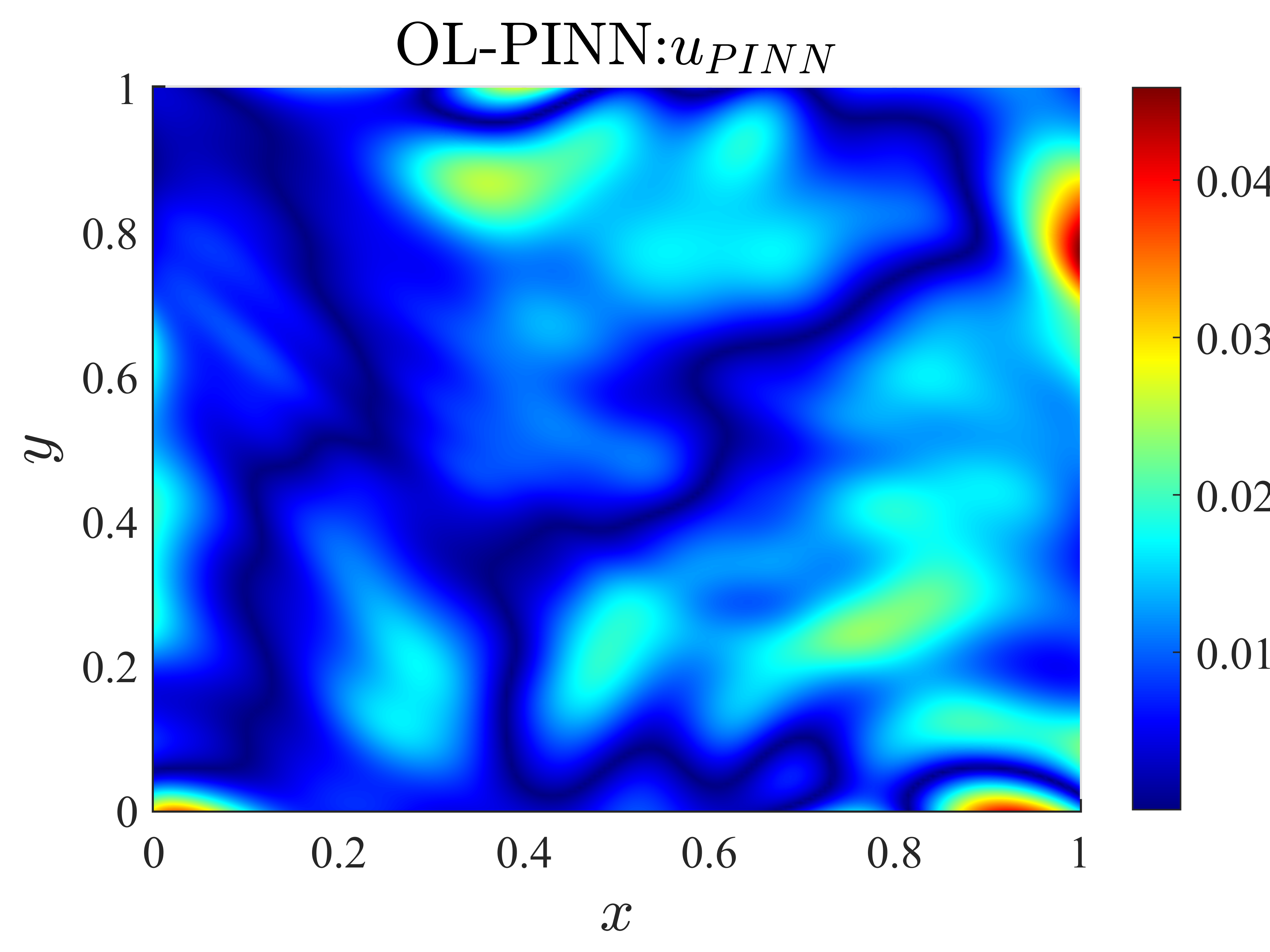}}

   \quad
		\subfloat[$|u_{ref} - u_{Op}|$ \label{fig8:e}]{
			
			\includegraphics[scale=0.25]{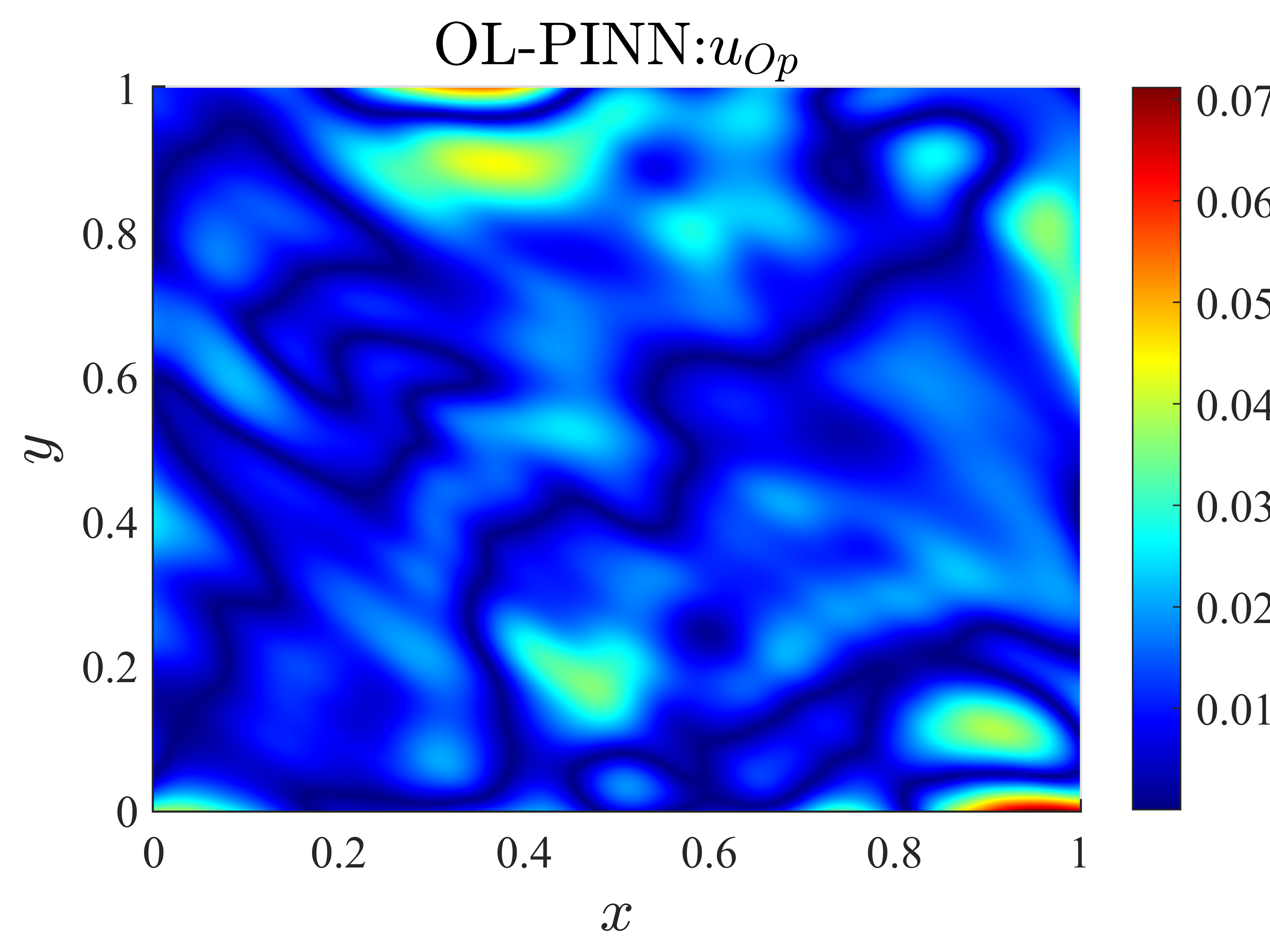}}
   \subfloat[$|u_{ref} - u_{PINN}|$  using  partial BCs\label{fig8:f}]{
			
			\includegraphics[scale=0.25]{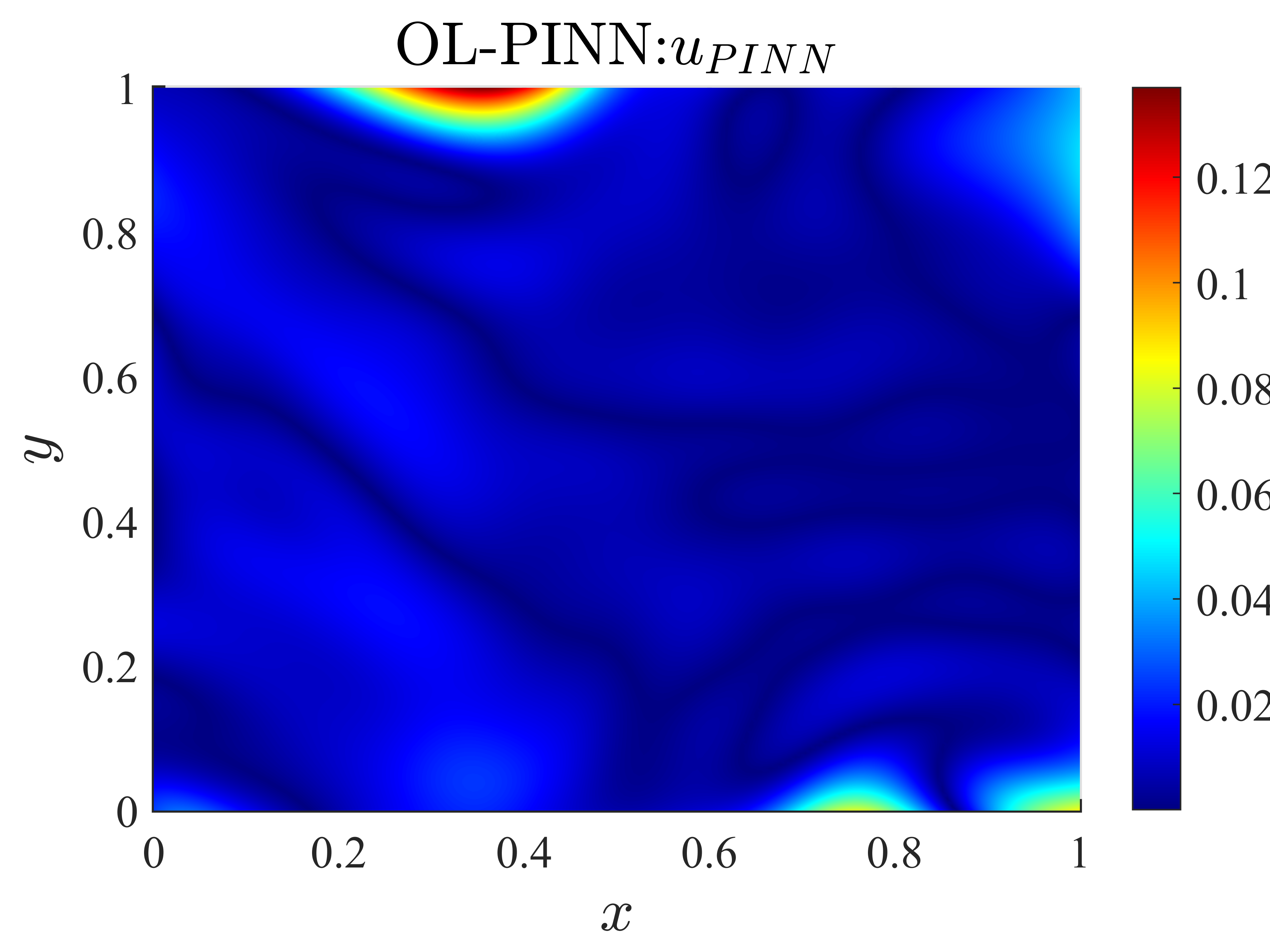}}
		\subfloat[Loss vs number of epochs\label{fig8:g}]{
			
			\includegraphics[scale=0.25]{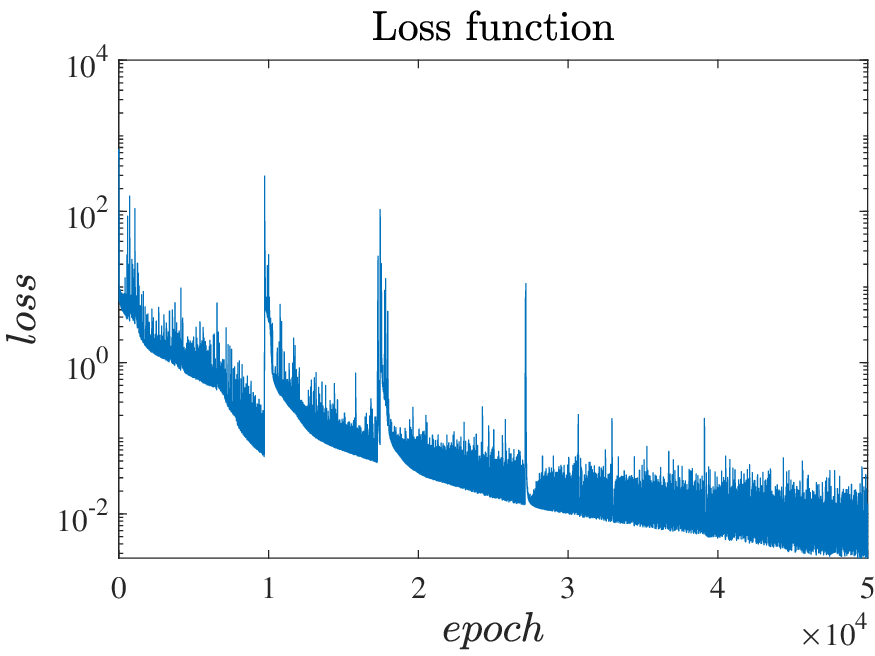}}
		\subfloat[Snapshot along $y =0.6$\label{fig8:h}]{
			
			\includegraphics[scale=0.25]{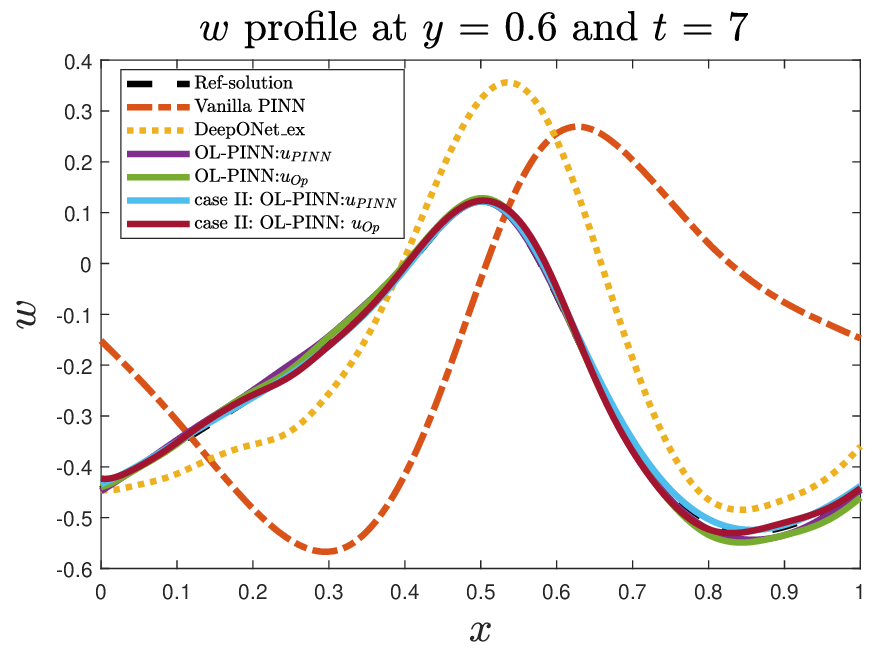}}

		\caption{Example 4: (a) Reference solution. (b)-(f) The absolute error between the reference solution and the prediction with different models. (g) Loss vs Epoch. (h) Comparison of the snapshot along $y=0.6$. Here we use the spatial-temporal resolution $22\times 22 \times 20$ as the set of residual points.}
		\label{ns_ex_error}
	\end{figure}

We show the reference solution at time $t=7$ as well as the corresponding absolute error for each model in Fig. \ref{fig8:a}-\ref{fig8:f}. The loss is given in Fig. \ref{fig8:g}, and the snapshot along $y=0.6$ is given in Fig. \ref{fig8:h}. 
Therefore, we can conclude that the present method enhances significantly the accuracy and efficiency of PINNs, and provides a very effective tool, especially for solving complex problems.

\begin{table}[htbp]
    \centering
    \begin{tabular}{|c|c|c|c|c|}
 \hline {Model } & Vanilla PINN& DeepONet\_ex& OL-PINN:$u_{PINN}$& OL-PINN:$u_{Op}$ \\
\hline {Case I} &  $30.16\pm71.60$ & $ 27.73\pm8.65 $ & \bm{$2.21 \pm 0.73 $} & $2.86\pm0.65$ \\

\hline {Case II} & $ 43.28\pm16.52 $ & $ 27.73\pm8.65 $ & \bm{$2.93 \pm 1.29 $} & $3.36\pm1.05$ \\

\hline
\end{tabular}
    \caption{Example 4: Mean and standard deviation of the relative $L^2(\%)$ test error for each model.}
    \label{table_ns}
\end{table}

\section{Conclusion}\label{sec:summary}
We developed in this work an Operator Learning Enhanced Physics-Informed Neural Network (OL-PINN) approach specifically tailored to effectively handle problems featuring sharp solutions. 
To address the challenge posed by such sharp solution problems, we combine the PINN approach with the pre-trained DeepONet specifically designed for a related class of smooth problems. We subsequently apply the OL-PINN methodology to a variety of challenging equations, including the nonlinear diffusion-reaction equation, Burgers equation, Lid-driven cavity flow, and the Navier-Stokes equation in vorticity form.

The results demonstrate that our current method not only improves prediction accuracy but also ensures the stability of the training process. Furthermore, it has been established that OL-PINN exhibits exceptional generalization capabilities with a minimal number of residual points. Furthermore, our OL-PINN has shown promise as an effective solution for ill-posed problems, particularly those constrained by limited boundary conditions.


\section*{CRediT authorship contribution statement}

\textbf{Bin Lin}: Methodology, Investigation, Coding, Writing - original draft, Writing - review \& editing, Visualization.

\textbf{Zhiping Mao}: Conceptualization, Methodology, Investigation, Coding, Writing - original draft, Writing - review \& editing, Visualization, Supervision, Project administration, Funding acquisition.

\textbf{Zhicheng Wang}: Conceptualization, Methodology, Investigation, Writing - original draft, Writing - review \& editing, Supervision, Project administration.

\textbf{George Em Karniadakis}: Conceptualization, Methodology, Writing - original draft, Writing - review \& editing, Project administration.

\section*{Declaration of competing interest}
The authors declare that they have no known competing financial interests or personal relationships that could have appeared to influence the work reported in this paper.

\section*{Acknowledgment}
This work was supported by the National Key R\&D Program of China (Grant No. 2022YFA1004500) and the National Natural Science Foundation of China (Grant No. 12171404).

\bibliographystyle{elsarticle-num-names}
\bibliography{ref}

\appendix
\section{Parameters used in this work}\label{sec:apd:parameters}

In this appendix, we present some of the parameters unmentioned previously.

\begin{table}[htbp]
    \centering
    
    \begin{tabular}{lllll}
\hline & Branch net & {\makecell{Activation \\ function}} & Trunk net & {\makecell{Activation \\ function}} \\
\hline Section \ref{section:onedim}  & [6, 3] & tanh & [3, 3]& tanh \\

Section \ref{section:caseI} & [32, 8] & ReLU & [24 , 24 , 24, 8] & tanh\\
Section \ref{section:caseII} & [32, 8] & ReLU & [24 , 24 , 24, 8] & tanh \\
Section \ref{subsection:cavity}  & [32, 8] & ReLU & [24 , 24 , 24, 8] & tanh \\
Section \ref{subsection:vorticity}  & [256, 64] & ReLU & [192 , 192 , 192, 64] & tanh \\
\hline
\end{tabular}
\caption{The architecture of the (pretrianed) DeepONet for each problem.}
\label{pre_train}
\end{table}

\begin{table}[htbp]
    \centering
    
    \begin{tabular}{lllll}
\hline 
&\multicolumn{2}{c}{Branch net} &\multicolumn{2}{c}{Trunk net}
\\
    
\hline & Input$(f)$ & Dataset size & Input$(x,t)$ & Dataset size\\

\hline Section \ref{section:onedim}  &  $a\in[0,1]$ & $100 \times 51$ & $[-1,1]$& $100 \times 51$ \\

Section \ref{section:caseI} & $\nu \in [0.02/\pi,0.06/\pi]$ & $40\times50\times50$ & $[-1,1]\times[0,0.9]$ &$40\times161\times100$\\

Section \ref{section:caseII} & $\nu \in [0.005/\pi,0.01/\pi]$ & $100\times100\times201$ & $[-1,1]\times[0,0.6]$ & $100\times100\times201$ \\
Section \ref{subsection:cavity}  & $Re \in [400, 680]$ & $15\times50\times50$ & $[0,1]\times[0,1]$ & $15\times101\times101$ \\
Section \ref{subsection:vorticity}  & $w|_{(0,1)^2 \times(5,6]}$ & $450\times64\times64\times10$  & $[0,1]^2\times(6,7]$ & $450\times64\times64\times10$ \\
\hline
\end{tabular}
\caption{The range and the data size of the input for the (pretrianed) DeepONet for each problem.}
\label{pre_train}
\end{table}

\begin{table}[htbp]
    \centering
    \begin{tabular}{llll}
\hline & $x^c$ & $x^r$ & $x^{test}$ \\
\hline Section \ref{section:onedim} & 201 & 6& 81 \\
Section \ref{section:caseI}  & $201 \times 101$ & $51 \times 10$ & $111\times 91$\\
Section \ref{section:caseII}  & $201 \times 51$ & $21 \times 3$  & $301 \times 41$ \\
Section \ref{subsection:cavity} Case I & $201 \times 201$ & $101 \times 101$& $225 \times 225$ \\
Section \ref{subsection:cavity} Case II & $201 \times 201$ & $31 \times 31$ & $225 \times 225$ \\
Section \ref{subsection:cavity} Case II, partial BC & $201 \times 201$ & $101 \times 101$& $225 \times 225$ \\
Section \ref{subsection:vorticity} & $64 \times 64 \times 20$ & $22 \times 22 \times 20$  & $190 \times 190 \times 40$ \\
\hline
\end{tabular}
\caption{Size of the Data set used in OL-PINN for each case.}
\end{table}

\begin{table}[htbp]
    \centering
    
    \begin{tabular}{lllll}
\hline & Branch net & {\makecell{Activation \\ function}} & Trunk net & {\makecell{Activation \\ function}} \\
\hline Section \ref{section:onedim}  & [6, 3] & tanh & [3, 3]& tanh \\

Section \ref{section:caseI}$, \nu = \frac{0.001}{\pi}$ & [32, 8] & ReLU & [24 , 24 , 24, 8] & tanh\\
Section \ref{section:caseI}$, \nu = 0$ & [32, 8] & ReLU & [8 ,  8] & tanh\\
Section \ref{section:caseII} & [32, 8] & ReLU & [24 , 24 , 24, 8] & tanh \\
Section \ref{subsection:cavity}  & [1,1] & ReLU & [1,1] & tanh \\
Section \ref{subsection:vorticity}  & [256, 64] & ReLU & [192 , 192 , 192, 64] & tanh \\
\hline
\end{tabular}
\caption{The architecture of the NN for each case.}
\label{NN}
\end{table}

\begin{table}[htbp]
    \centering
    
    \begin{tabular}{lllll}
\hline 
&\multicolumn{2}{c}{Branch net} &\multicolumn{2}{c}{Trunk net}
\\
    
\hline & Input$(f)$ & Dataset size & Input$(\boldsymbol{x},\Tilde{u} )$ & Dataset size\\

\hline Section \ref{section:onedim}  &  $a=5,10$ & $1 \times 51$ & $([-1,1],\Tilde{u} )$& $-$ \\

Section \ref{section:caseI}, $\nu= \frac{0.001}{\pi}$& $\nu =\frac{0.001}{\pi}$ & $1 \times 16 \times 16$ & $([-1,1]\times[0,0.9],\Tilde{u} )$ &$-$\\

Section \ref{section:caseI}, $\nu=\frac{0.001}{\pi}$, no BCs  & $\nu =\frac{0.001}{\pi}$ & $1\times32 \times32$ & $([-1,1]\times[0,0.9],\Tilde{u} )$ &$-$\\

Section \ref{section:caseI}, $\nu=0$   & $\nu =0$ & $1\times4 \times4$ & $([-1,1]\times[0,0.9],\Tilde{u} )$ &$-$\\

Section \ref{section:caseII} & $\nu =0.008/\pi$ & $1 \times100\times201$ & $([-1,1]\times[0.6,0.8],\Tilde{u} )$ & $-$ \\

Section \ref{subsection:cavity}, Case I  & $Re = 1000$ & $1\times4\times4$ & $([0,1]\times[0,1],\Tilde{u} )$ & $-$ \\
Section \ref{subsection:cavity}, Case II  & $Re = 1000$ & $1\times2\times2$ & $([0,1]\times[0,1],\Tilde{u} )$ & $-$ \\

Section \ref{subsection:vorticity}  & $w|_{(0,1)^2 \times(5,6]}$ & $1\times64\times64\times10$  & $([0,1]^2\times(6,7],\Tilde{u} )$ & $-$ \\
\hline
\end{tabular}
\caption{The input of the NN for each case. Note that the data size of trunk network is determined by the number of collocation point of $u_{Op}$.}
\label{pre_train}
\end{table}

\begin{table}[htbp]
    \centering
    \begin{tabular}{llll}
\hline & $w_1$ & $w_2$ & $w_3$ \\
\hline Section \ref{section:onedim} & $1$ & $1$& 200 \\
Section \ref{section:caseI}$, \nu = 0.001/\pi$  & $5$ & $5$ & $50$\\
Section \ref{section:caseI}$, \nu = 0$  & $1$ & $1$ & $10$\\
Section \ref{section:caseII}  & $1$ & $1$  & $100$ \\
Section \ref{subsection:cavity} Case I & $70$ & $70$& $50$ \\
Section \ref{subsection:cavity} Case II, first 50000 epoch  & $1$ & $1$ & $1$ \\
Section \ref{subsection:cavity} Case II: last 30000 epoch  & $200$ & $200$ & $30$ \\
Section \ref{subsection:vorticity} Case I & $100$ & $100$  & $50$ \\
Section \ref{subsection:vorticity} Case II, partial BC & $100$ & $100$  & $20$ \\
\hline
\end{tabular}
\caption{The weights in \eqref{loss:ol-pinn} for each case.}
\end{table}

\begin{table}[htbp]
    \centering

    \begin{tabular}{lll}
\hline &  PINN &  {\makecell{Activation \\ function}} \\
\hline Section \ref{section:onedim}  & Depth 2 \& Width 128 & tanh\\
Section \ref{section:caseI}   & Depth 3 \& Width 128 & tanh \\
Section \ref{section:caseII}   & Depth 3 \& Width 128 & tanh\\
Section \ref{subsection:cavity}  & Depth 4 \& Width 200 & tanh\\
Section \ref{subsection:vorticity} & Depth 5 \& Width 96 & tanh\\
\hline
\end{tabular}

\caption{The architecture of the PINN for each case.}
\end{table}

	
	

\end{document}